\documentclass[acmsmall]{acmart}

\usepackage{multirow}
\usepackage{graphicx}
\usepackage{xcolor}
\definecolor{BLACK}{named}{black}
\newcommand{\revised}[1]{\textcolor{black}{#1}}
\AtBeginDocument{%
  }

\setcopyright{acmlicensed}
\copyrightyear{2018}
\acmYear{2018}
\acmDOI{XXXXXXX.XXXXXXX}

\acmJournal{JACM}
\acmVolume{37}
\acmNumber{4}
\acmArticle{111}
\acmMonth{8}

\begin{document}

\title{InstructTime++: Time Series Classification with Multimodal Language Modeling via Implicit Feature Enhancement}

\author{\revised{Mingyue Cheng}}
\author{\revised{Xiaoyu Tao}}
\author{\revised{Huajian Zhang}}
\author{\revised{Qi Liu}}
\authornote{Correspondence to: Qi Liu <qiliuql@ustc.edu.cn>}
\author{\revised{Zhiding Liu}}
\author{\revised{Yucong Luo}}
\author{\revised{Yiheng Chen}}
\author{\revised{Enhong Chen}}

\affiliation{
  \institution{State Key Laboratory of Cognitive Intelligence, University of Science and Technology of China}
  \city{Hefei}
  \state{Anhui}
  \country{China}
}
\email{mycheng@ustc.edu.cn}
\email{txytiny@mail.ustc.edu.cn}
\email{zhjustc@mail.ustc.edu.cn}
\email{qiliuql@ustc.edu.cn}
\email{cheneh@ustc.edu.cn}

\renewcommand{\shortauthors}{Mingyue Cheng et al.}

\begin{abstract}
Most existing time series classification methods follow a discriminative modeling paradigm that directly maps input sequences to one-hot encoded class labels. Although effective, this paradigm suffers from two inherent limitations: it is difficult to effectively exploit contextual features, and one-hot label distribution fails to capture the semantic relationships among classes. In this paper, we propose a novel framework, termed InstructTime, which reformulates time series classification as a multimodal generative task. The core idea is to treat continuous numerical sequences, discrete textual features, and task-specific instructions as multimodal inputs, while representing class labels as textual outputs, with classification realized through the generative capability of tuned language models (LMs). To bridge the modality gap, we introduce a time series discretization module that converts continuous numerical sequences into discrete temporal tokens, thereby alleviating the inconsistency between numerical and textual modalities. In addition, we employ an alignment projection layer together with a generative self-supervised  pre-training strategy to enhance cross-modal representation alignment. Lastly, we further fine-tune the aligned language model with task-specific instructions to strengthen its multimodal reasoning capability. While InstructTime effectively models explicit contextual features, language models are generally not well-suited to capturing implicit features that are not directly observable from raw time series or contextual features, such as latent temporal and structural patterns. Based on the above analysis, we extend InstructTime to InstructTime++ by introducing an improved multimodal generative framework that additionally captures implicit features to compensate for the limited inductive bias of current LMs-based time series classifiers. In particular, InstructTime++ leverages a collection of specialized toolkits to automatically mine informative implicit patterns from multiple views of raw time series and contextual inputs. In this work, we instantiate this framework from two complementary perspectives by designing statistical feature extraction and vision–language–based image captioning for textual grounding. The discovered implicit features are subsequently translated into textual descriptions, enabling seamless integration with the original InstructTime framework. Finally, extensive experiments on multiple benchmark datasets demonstrate the superior performance of InstructTime++. \revised{This work further paves the way for agentic time series classification}\footnote{Our code is at https://github.com/Mingyue-Cheng/InstructTime.}. 

\end{abstract}



\received{20 February 2007}
\received[revised]{12 March 2009}
\received[accepted]{5 June 2009}

\maketitle
\section{Introduction}
\begin{figure*}[ht]
	\centering
	\includegraphics[width=1.0\textwidth]{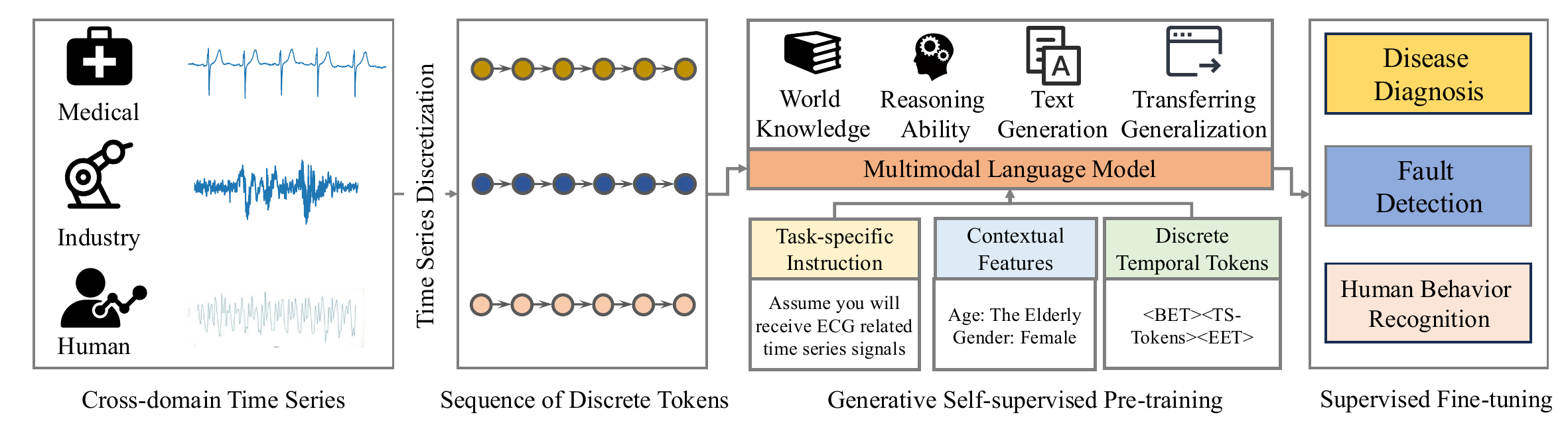}
	\caption{Illustrating the overall pipeline of proposed InstructTime.}
	\label{fig:example}
\end{figure*}
Time series classification (TSC) is a crucial task in data science~\cite{cheng2023formertime}, driven by increasing demand across diverse applications, such as healthcare diagnosis~\cite{zheng2014time}, industrial anomaly detection~\cite{chen2025prospective}, and human activity recognition~\cite{liu2019ekt}. TSC aims to assign a given time series to predefined labels based on its temporal patterns and contextual features \revised{~\cite{liu2024generative}}.

Over the past decades, the task of TSC attracts extensive research attention, leading to various classical approaches~\cite{gupta2020approaches}, including distance-based~\cite{petitjean2014dynamic}, shapelet-based~\cite{ye2009time}, dictionary-based~\cite{lin2012rotation}, and feature-based~\cite{deng2013time} methods. Despite their effectiveness, the former three categories often suffer from high computational costs and limited scalability, while the last category typically relies on labor-intensive handcrafted efforts during feature engineering. Recent advances further shift TSC toward deep learning–based approaches, with representative network architectures such as convolutional neural networks (CNNs)~\cite{cheng2024convtimenet}, graph neural networks (GNNs)~\cite{han2021dynamic}, and self-attention-based Transformers~\cite{cheng2026timemae}. These approaches exhibit a strong capacity to effectively capture complex temporal dependencies via automatically learning feature representations ~\cite{wang2024tabletime}. In terms of optimization strategies, it is common to adopt a discriminative learning paradigm, which minimizes the discrepancy (e.g., the cross-entropy loss) between the predicted distribution and the target one-hot distribution~\cite{dempster2021minirocket}.

While the above modeling paradigm is effective, it suffers from two inherent limitations. First, these previous deep learning–based networks often struggle to effectively incorporate contextual features, which are typically provided in textual form~\cite{rendle2010factorization,cao2023tempo}. For example, in terms of medical diagnosis, incorporating textual information such as patient age and gender plays an important role in an electrocardiogram (ECG) ~\cite{anguita2013public} analysis. Second, the widely used discriminative optimization strategies often use one-hot distribution to represent the class label, which struggle to reflect the semantic relationships among classes~\cite{hinton2015distilling,li2025incomplete}. Actually, in domains such as human activity recognition (HAR)~\cite{cheng2025cross}, activities like walking and jogging are semantically more similar to each other than to lying down. Based on the above analysis, a new modeling paradigm is required that can naturally incorporate contextual features and capture semantic relationships among classes.

In response to the aforementioned limitations, we propose a novel modeling paradigm that reformulates time series classification as a multimodal generative task. By representing class labels as textual descriptions rather than one-hot distributions, this paradigm enables the modeling of semantic relationships among classes while naturally facilitating the incorporation of contextual features expressed in textual form. As illustrated in Figure~\ref{fig:example}, inspired by recent advances in language models~\cite{brown2020language,radford2018improving}, our approach jointly considers continuous numerical sequences, discrete textual features, and task-specific instructions as multimodal inputs, and performs classification via textual label generation.
However, directly adopting LMs for TSC introduces substantial challenges due to the inherent inconsistency between numerical and textual modalities. 

To address the above challenge, we propose InstructTime, which introduces a time series discretization strategy based on vector-quantized (VQ) networks~\cite{van2017neural}  that transform continuous numerical sequences into discrete temporal tokens, thereby alleviating the inconsistency between numerical and textual modalities.
Beyond discretization, we further incorporate an alignment projection layer implemented via a multilayer perceptron (MLP) ~\cite{liu2021pay} together with a generative self-supervised  pre-training strategy to enhance cross-modal representation alignment. Finally, we perform task-specific instruction fine-tuning on the aligned language model, enabling it to better leverage task guidance and improve its multimodal reasoning capability for TSC. Based on the above designs, InstructTime enables a paradigm shift in TSC, moving from discriminative mapping to generative multimodal reasoning.

Although InstructTime effectively exploits explicit contextual features, it remains limited in modeling implicit features that are not directly observable from raw time series or contextual inputs. Such implicit features, such as latent temporal dynamics and underlying structural patterns, play a critical role in time series classification but are not naturally captured by LMs–based classifiers. This limitation can be attributed to the lack of temporal inductive bias in existing LMs. Motivated by this observation, we further extend InstructTime to InstructTime++ by introducing an improved multimodal generative framework, enhanced with implicit feature modeling. The core idea is to extend the original framework by introducing modules that can automatically mine informative implicit patterns from multiple views of raw time series and contextual inputs. In particular, InstructTime++ leverages a collection of specialized toolkits to extract complementary forms of implicit features. Specifically, we instantiate this paradigm from two complementary perspectives to induce informative implicit features. To this end, we design a collection of predefined tools to capture statistical features, and vision–language–based image captioning to extract visual features by grounding time series visualizations. The discovered implicit features are subsequently translated into textual descriptions, enabling seamless integration with the original InstructTime framework. By incorporating  implicit features, InstructTime++ effectively alleviates the inherent limitations of LMs in capturing temporal dynamics and structural patterns, thereby enhancing their multimodal reasoning capability in time series classification. Extensive experimental results across multiple benchmark datasets demonstrate the effectiveness of InstructTime++.
\section{Preliminaries}
In this section, we formally define the time series classification problem and introduce the related learning settings considered in this work.

\subsection{Problem Definition}
\paragraph{Time Series Classification.}
Given a dataset $\mathcal{D} = \{(\mathbf{x}_i, y_i)\}_{i=1}^{N}$, where each $\mathbf{x}_i \in \mathbb{R}^{L \times H}$ is a time series with $L$ time steps and $H$ channels, and $y_i \in \mathcal{C} = \{c_1, c_2, \ldots, c_K\}$ is the corresponding label, the goal of TSC is to learn a classification function
$
g : \mathbb{R}^{L \times H} \rightarrow \mathcal{C},
$
which accurately predicts the discrete label for a given time series. This task requires capturing the underlying temporal patterns.

\paragraph{Multimodal Generative Time Series Classification.}
Given a dataset
$
\mathcal{D} = \{(\mathbf{x}_i, \mathbf{c}_i, y_i)\}_{i=1}^{N},
$
where $\mathbf{x}_i \in \mathbb{R}^{L \times H}$ denotes a time series, \revised{$\mathbf{c}_i$ represents the associated contextual features, and $y_i \in \mathcal{C}$ is the corresponding label, the goal of multimodal generative time series classification is to learn a generative function
$
g : (\mathbb{R}^{L \times H}, \mathbf{c}) \rightarrow \mathcal{Y},
$
which performs classification by generating textual outputs conditioned on multimodal inputs.}
The class label is then determined by identifying predefined class-specific keywords from the generated text.

\paragraph{Cross-domain Time Series Representation.}
Consider $N$ distinct domains $\mathcal{D}_1, \mathcal{D}_2, \ldots, \mathcal{D}_N$, where each domain $\mathcal{D}_i$ contains $M_i$ samples $\{\mathbf{x}_1^{i}, \mathbf{x}_2^{i}, \ldots, \mathbf{x}_{M_i}^{i}\}$, and \revised{each $\mathbf{x}_j^{i} \in \mathbb{R}^{L \times H}$ represents a time series.} Cross-domain representation learning aims to learn a function
$
\phi : \mathbb{R}^{L \times H} \rightarrow \mathbb{R}^{d},
$
which transforms any time series into a $d$-dimensional representation. Such domain-invariant representations facilitate effective knowledge transfer and improved generalization across datasets.

\paragraph{Domain-specific Time Series Classification.}
After obtaining a domain-invariant representation, the task in a target domain is to classify each \revised{transformed time series
$
\mathbf{z}_j^{i} = \phi(\mathbf{x}_j^{i}) \in \mathbb{R}^{d}
$
into one of $C$ predefined labels.} Formally, a classification function
$
h : \mathbb{R}^{d} \rightarrow \mathcal{C}
$
assigns a label $y_j^{i} \in \mathcal{C}$ to each representation. The objective is to achieve high classification accuracy in the target domain.

\section{InstructTime}
In this section, we introduce InstructTime, a novel approach that reformulates time series classification as a multimodal generation task. By treating continuous numerical sequences, discrete textual features, and task-specific instructions as distinct modalities, InstructTime unifies classification tasks within an instruction-driven generative framework.
\begin{figure*}[ht]
	\centering
	\includegraphics[width=1.0\textwidth]{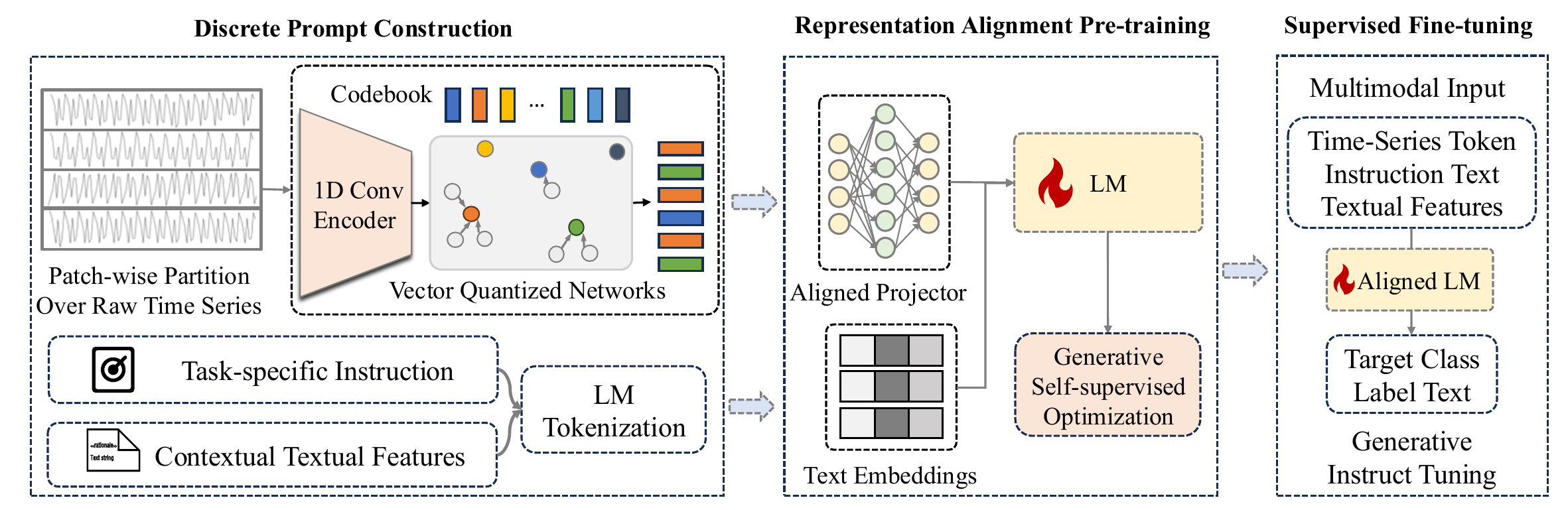}
    \caption{Illustration of the overall framework of the proposed InstructTime. }
	\label{fig:framework}
\end{figure*}
\subsection{Framework Overview}
Figure~\ref{fig:framework} illustrates the overall framework of InstructTime, which consists of three key phases. The process begins with a patch-wise partitioning of raw time series, encoded by VQ networks to obtain discrete temporal tokens. Task-specific instructions are incorporated to enrich these tokens with contextual features. The sequences are tokenized by the alignment projector and aligned with textual embeddings, preparing the inputs for the LM. 
Subsequently, the LM is optimized in a generative  self-supervised  manner across multiple domains. Finally, supervised fine-tuning (SFT) with multimodal inputs, including discrete temporal tokens, instruction text, and textual features, adapts the aligned LM to the downstream task via generative instruct tuning.

\subsection{Discrete Prompt Construction}
\subsubsection{Time Series Discretization} 
Time series differ from text in that they consist of continuous numerical sequences rather than discrete linguistic units. This distinction introduces unique challenges for effective modeling. On one hand, the semantic information contained in individual time steps is relatively sparse ~\cite{cheng2023formertime}. Unlike words in a sentence that carry rich and distinct meanings, numerical values generally become informative only when aggregated into temporal patterns. On the other hand, the length of time series can be considerably extensive~\cite{zhou2021informer}, often spanning thousands or millions of data points to represent comprehensive temporal dynamics. These characteristics hinder the direct application of language models to raw time series.

Directly applying language models to time series data without accounting for these intrinsic differences can lead to significant performance degradation, underscoring the need for specialized adaptations capable of handling the unique characteristics of time series. In this work, we decide to adopt a simple and direct idea of time series discretization. Although several related approaches are proposed, such as the well-known symbolic aggregate approximation (SAX)~\cite{lin2007experiencing}, we argue that traditional discretization techniques face notable limitations in practical applications.  A primary concern is the substantial information loss during the compression process, where the dimensionality reduction inherent in these methods often leads to the omission of critical nuances and patterns within the time series. This compression loss can significantly affect the subsequent analysis, leading to less accurate results. 

Based on the above analysis, we design a novel technique for time series discretization. Specifically, we adopt VQ networks~\cite{van2017neural,gray1984vector, gao2023adaptive} to convert continuous subsequences into discrete representations. The core idea is to assign each local sub-series an identity code selected from a predefined codebook, where the assignment is learned through a reconstruction-based optimization process within an auto-encoder architecture. In this work, we employ the temporal convolutional network (TCN)~\cite{oord2016wavenet} as the backbone for both the encoder and the decoder. In general, the correspondence between a sub-series and candidate codes is measured using a classical similarity metric, such as squared Euclidean distance, and the goal is to identify the discrete sequence that minimizes the cumulative reconstruction distance.

Formally, for the $i$-th domain, we assign a trainable embedding matrix denoted by $E_i =\{e_k^i\}_{k=1}^K = \{e_1^i, e_2^i., \ldots,  e_K^i\}$, where each $e_k^i$ represents one of the $K$ embedding vectors. The indices associated with these continuous vectors serve as the corresponding discrete codes. Before feeding the raw time series into the encoder of the VQ network, we first apply a patch embedding operation. \revised{Given a time series instance $X_{ij} = \{z_1, z_2, \ldots, z_L \}$ from the $i$-th domain and the $j$-th example, it is segmented into $P$ non-overlapping contiguous patches $(s_1, s_2, \ldots, s_P)$.} Each patch $s_p$ corresponds to a subsequence of $z_{1:T}$ containing $|s_p|$ successive time points. By applying a weight-sharing 1D convolutional operation, each segment $s_p$ is transformed into a representative vector.

\revised{In the forward pass of the VQ network, a sequence of latent representations $(\mathbf{z}_1, \mathbf{z}_2, \ldots, \mathbf{z}_P)$ is discretized by mapping each $\mathbf{z}_p$ to its nearest neighbor in the codebook. Specifically, each $\mathbf{z}_p$ is replaced with $\mathbf{e}_k$, where
$
k = \arg\min_{k} \| \mathbf{z}_p - \mathbf{e}_k \|^2.
$
The output of this layer is the quantized sequence $(\hat{\mathbf{z}}_1, \hat{\mathbf{z}}_2, \ldots, \hat{\mathbf{z}}_P)$.
Here, $\mathbf{z}_p$ denotes the continuous latent representation produced by the encoder, while $\hat{\mathbf{z}}_p$ denotes the corresponding quantized representation obtained from the codebook.
For example, consider a segment $\mathbf{s}_1 = \mathbf{z}_{1:10}$. If this segment is assigned to the code indexed by $k = 12$, then its quantized representation is given by $\hat{\mathbf{z}}_1 = \mathbf{e}_{12}$.}

To train the entire VQ network—including the encoder, decoder, and the codebook embeddings—we optimize a reconstruction loss. Formally, the overall objective can be expressed as:
\begin{equation}
	\label{loss}
	\mathcal{L} = \text{log} p(x|z_q(x)) + ||\text{sg}[z_e(x)] - e||^2_2  + \beta ||z_e(x) - \text{sg} [e]||^2_2,	
\end{equation}

where $\text{sg}$ denotes the stop-gradient operator, which behaves as an identity function during the forward pass but yields zero partial derivatives during backpropagation, effectively treating its operand as a constant. This loss function contains three components, each responsible for training different parts of the VQ network. The first term is the reconstruction loss, which updates the encoder and decoder. Besides reconstruction, the loss includes two additional terms. Because the $\arg\min$ operator in the forward pass is non-differentiable, gradients are approximated using the straight-through estimator (STE)~\cite{bengio2013estimating} during the backward pass. As a result, the codebook is updated using an exponential moving average of the continuous encoder outputs, as reflected in the second term of Equation~\ref{loss}. Finally, since the embedding space is dimensionless and may otherwise grow arbitrarily when the encoder updates faster than the codebook, a third component—the commitment loss—is introduced to ensure that the encoder commits to a selected embedding and prevents its output magnitudes from diverging.
\begin{figure}[h]
	\centering
	\includegraphics[width=0.65\textwidth]{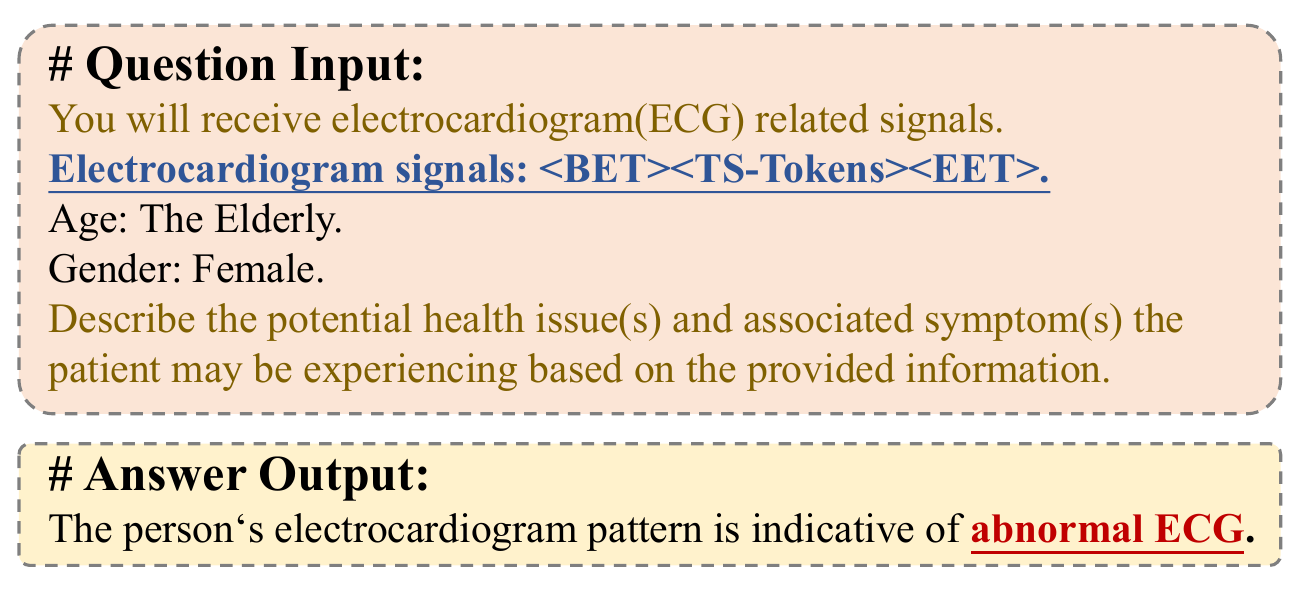}
	\caption{ \revised{Illustration of the proposed prompt template, including task-specific instruction, contextual features, and discrete temporal tokens used for label generation.}}
	\label{fig:prompt}
\end{figure}
\subsubsection{Prompt Template Design of InstructTime}
For each domain, we utilize fixed prompt templates for target label generation. Ultimately, our fixed prompt template is structured as shown in Figure~\ref{fig:prompt}. This template totally contains three sides of information: domain-specific instruction, contextual features, and discrete temporal tokens.  Particularly, \revised{``TS-Tokens''} denotes the sequence of discrete time series tokens projected by VQ networks, while ``<BET>'' and ``<EET>'' denote the placeholders. It should be noted that prompt templates play a vital role in the language generation task. In this work, we highlight the following three guidelines in designing a prompt template: (1) inclusion of candidate label information: we incorporate candidate label information within the prompt to provide explicit cues to the LM. This approach primes the LM to consider these labels as part of the solution space, thereby enhancing its ability to associate given inputs with potential outcomes. (2) Consistency in sequence length: given the multi-domain nature of our task, it is crucial to maintain uniform sequence lengths for time series across different domains. This consistency ensures that the LM is not biased by varying lengths and can learn domain-invariant features, facilitating better generalization and classification. (3) Completion of predictive answer output: instead of merely predicting discrete answers, we instruct the model to generate a complete sentence.

\subsection{Hybrid Encoding  Strategies}
Considering the significant representation gap between the modalities of temporal tokens and textual words, we implement a hybrid encoding strategy to convert the input prompt into latent vectors, namely embeddings. As illustrated in \autoref{fig:framework}, we process all textual content using the LM’s native tokenization and embedding mechanisms to obtain token representations and their corresponding embeddings.
During prompt construction, it is crucial to ensure that embeddings from both modalities are properly aligned to support effective multimodal reasoning. Therefore, we additionally develop an aligned projector module to mitigate the potential representation gap. Specifically, the aligned projector is fine-tuned to convert discrete temporal tokens into a format that is coherent with the embeddings used by the LM. This alignment is facilitated through a training strategy that integrates the temporal and textual embeddings within a unified latent space. For practical implementation, a straightforward MLP can be employed for this purpose.
Once the input prompt is converted into a sequence of embeddings, the LM generates target labels. The LM employs its pre-trained neural network architecture to process the sequence, utilizing the contextual information encoded within the embeddings to inform its predictions. The embedding sequence is processed through multiple layers of the LM, progressively refining the representations before reaching the output layer. Here, the LM integrates multimodal embeddings and generates the final prediction through autoregressive decoding. This prediction is based on both the time series and the accompanying contextual features, enabling more accurate time series classification.

\subsection{Training Procedure}
\label{training_phase}
We now present the training strategy for the model parameters. In our work, we adopt full fine-tuning strategies during training of the LM. Functionally speaking, there exist two key training phases: (1) generative self-supervised  pre-training across domains; and (2) supervised fine-tuning within a domain.  
During the first training phase, the generative self-supervised  pre-training~\cite{radford2018improving} is designed to further align the representation between discrete temporal tokens and language words. This is achieved through cross-domain generative pre-training, which serves a dual purpose. Firstly, it enhances the alignment of representations, ensuring that the temporal tokens are embedded in a manner that is cognitively coherent with the language model's understanding of natural language words. Secondly, through cross-domain self-supervised pre-training, the model's domain generalizability is significantly improved. This phase leverages the diversity of data across domains to teach the model to recognize and adapt to underlying patterns and structures, thus enriching the model's ability to handle domain-specific nuances during the subsequent fine-tuning phase.
In the second phase, domain-specific supervised generative fine-tuning~\cite{ouyang2022training} is employed to tailor the aligned LM to the target domain effectively. For this specific input prompt, we do not compute the regression loss, as it is used solely for domain alignment rather than task-specific supervision. Subsequently, the model is fine-tuned on a target domain using labeled data that captures the characteristics of the downstream task. This phase is critical as it allows the model to apply the generalized knowledge acquired during the pre-training phase to the specific target domain, enhancing its predictive performance. The fine-tuning adjusts the model parameters to optimize for domain-specific features, leading to a more refined model that is better equipped to handle the intricacies and particularities of the target domain's time series.

\section{InstructTime++}
In this section, we introduce InstructTime++, an extended framework that augments InstructTime with generative structural feature modeling for time series classification. We first outline the overall framework of InstructTime++, followed by a detailed description of the implicit feature extraction module. Finally, we present the prompt template design used to integrate induced implicit features with explicit contextual features for effective multimodal reasoning.
\begin{figure*}[ht]
    \centering
    \includegraphics[width=1\linewidth]{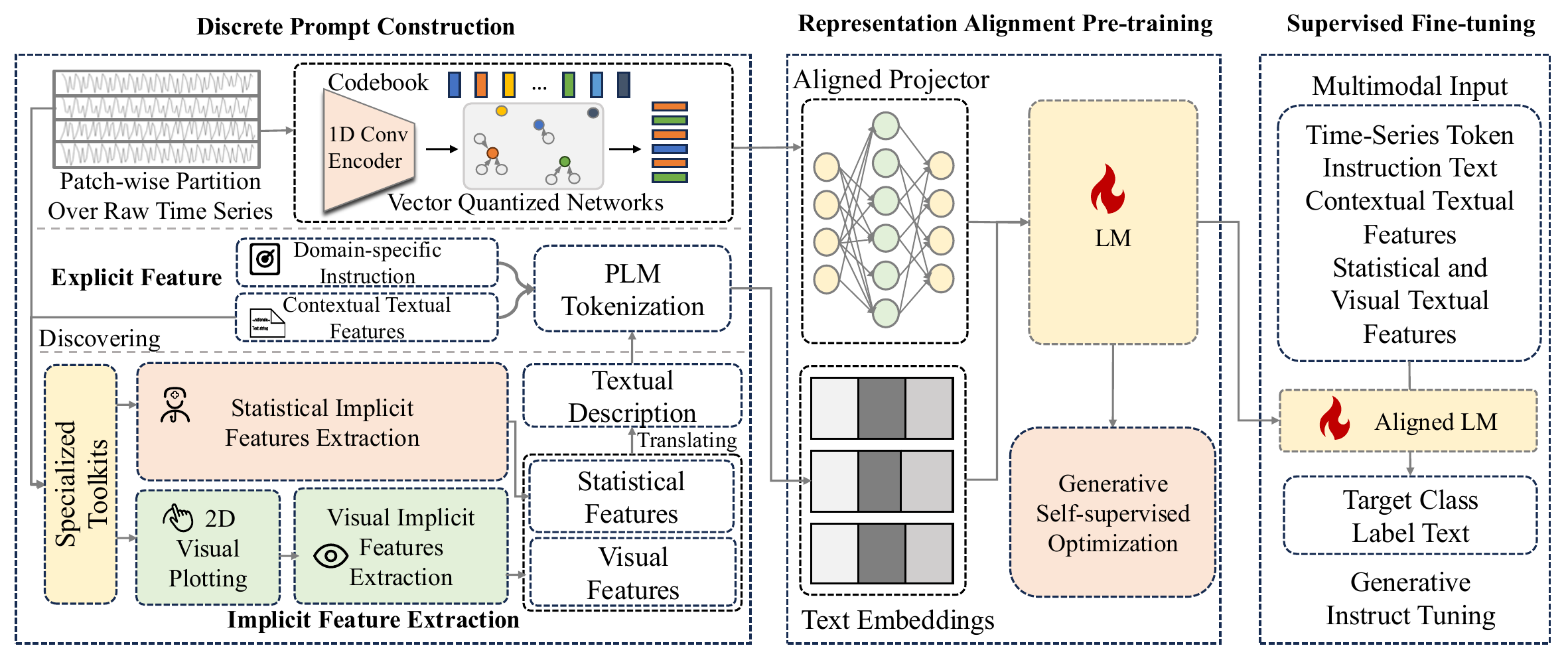}
    \caption{Illustration of the framework of extended InstructTime++.}
    \label{fig:Visualization}
\end{figure*}
\subsection{Framework Overview}
Figure~\ref{fig:Visualization} provides an overview of the InstructTime++ pipeline, illustrating how explicit and implicit features are progressively transformed and integrated for generative time series classification.
The framework begins by partitioning raw time series into patches and discretizing them into temporal tokens via VQ networks, making numerical sequences compatible with textual representations. In parallel, explicit contextual features and task-specific instructions are incorporated in textual form. To complement these explicit inputs, InstructTime++ further introduces an implicit feature extraction process, where statistical and visual patterns are mined using specialized toolkits and translated into natural language descriptions.
All textualized representations are then projected into a shared semantic space through an alignment projection layer and optimized under a generative self-supervised  objective to enhance cross-modal alignment. Finally, the aligned language model is fine-tuned with task-specific instructions, taking multimodal inputs and performing classification via generative prediction of label texts.

\subsection{Implicit Feature Extraction}
\subsubsection{Specialized Toolkits Construction}
Language models exhibit strong capability in modeling linguistic semantics, with inductive biases shaped by large-scale natural language corpora. Such corpora primarily capture syntactic and semantic regularities, while providing limited inductive support for numerical and temporal structures. As a result, general-purpose language models may not fully capture temporal dynamics and structural patterns that are critical for time series understanding. Even after discretization, numerical sequences are typically treated as symbolic tokens, without explicitly modeling temporal behaviors such as trends and periodicity. This limitation suggests that relying solely on end-to-end generative modeling may be insufficient to fully exploit implicit temporal information.
\revised{To address this limitation, InstructTime++ introduces a set of specialized toolkits for implicit feature construction. These toolkits aim to extract complementary temporal and structural patterns that are not directly observable from raw time series or explicit contextual features. Importantly, the proposed toolkits should be viewed as a flexible instantiation of the framework rather than a fixed design. As illustrated in Figure~\ref{fig:Visualization}, they are organized from two complementary perspectives, enabling a more comprehensive characterization of implicit temporal and structural patterns. Moreover, the modular design allows alternative feature extraction strategies to be incorporated without modifying the overall framework.}

\paragraph{Statistical Implicit Feature Extraction}
To compensate for statistical information introduced by explicit representations, we incorporate a rule-driven statistical feature extraction module that captures deterministic and interpretable statistical properties through predefined functions. Given a continuous time series $\mathbf{X}$, we construct a set of statistical functions $\mathbf{F}$, where each function is designed to characterize a statistical attribute. 
Specifically, the proposed module models the time series from three complementary perspectives. First, distributional statistics, including the mean, variance, skewness, and kurtosis, are computed to describe the underlying value distribution of $\mathbf{X}$. Second, to quantify the structural complexity and irregularity of temporal dynamics, complexity measures such as sample entropy and approximate entropy are employed to characterize the degree of unpredictability in the sequence evolution. Third, long-term temporal behaviors are explicitly modeled through linear trend estimation and periodicity detection, capturing global temporal dynamics.
All statistical functions in $\mathbf{F}$ are applied to $\mathbf{X}$ to produce a structured set of statistical features, which serve as implicit numerical representations.

\paragraph{Visual Implicit Feature Extraction}
While statistical features effectively capture numerical properties, they are often insufficient for characterizing the overall shape and morphological structure of time series, especially complex nonlinear hierarchical patterns. To solve the problem, we further leverage the visual understanding capability of VLMs to extract perception-driven structural patterns. Explicit contextual features guide the visual rendering strategy and the perceptual focus of the VLM, enabling the extraction of structurally relevant patterns aligned with the target task.
\revised{Specifically, a one-dimensional time series $\mathbf{X}$ is first transformed into a two-dimensional visualization $\mathbf{V}$ according to domain-informed rendering rules. The resulting image is then processed by a VLM to perceive structural characteristics at multiple temporal scales, including global trends, local shape patterns, and fine-grained fluctuations. These visual cues are aggregated to form perception-driven visual features that reflect the overall temporal morphology. 
From an efficiency perspective, visual rendering and VLM-based processing introduce additional computational cost compared to statistical feature extraction~\cite{song2025comprehensive}. However, these components can be optionally enabled based on task requirements. In practice, the resulting visual features can also be pre-computed and stored as textualized representations, thereby avoiding repeated overhead during inference. This design enables a flexible trade-off between representation richness and computational efficiency, while maintaining the modularity of the proposed framework.}

\begin{figure*}
    \centering
    \includegraphics[width=1\linewidth]{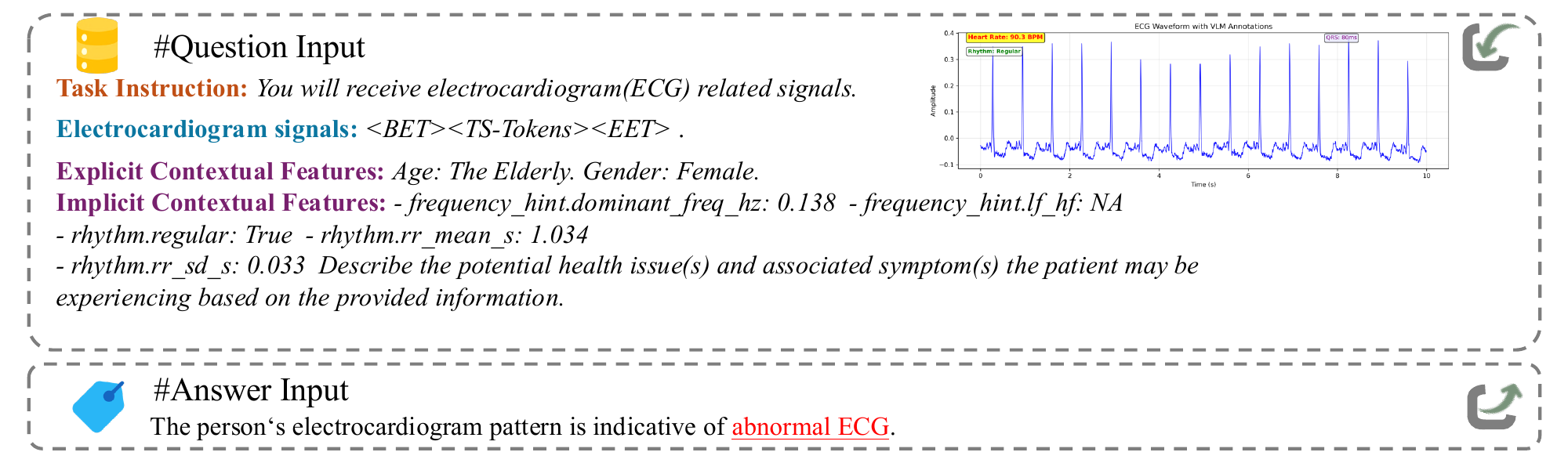}
    \caption{\revised{Illustration of the extended prompt template in InstructTime++, incorporating task-specific instructions, explicit contextual features, and implicit feature descriptions.}}
    \label{fig:Visualization}
\end{figure*}
\subsection{Implicit Feature Enhancement}
\subsubsection{Implicit Feature Representation}
Since the extracted implicit features originate from heterogeneous sources, such as statistical analysis results and visual perception signals, they naturally exist in different representation spaces and cannot be directly utilized by language models. Some existing approaches attempt to embed these features as vectors and feed them into language models ~\cite{tao2024hierarchical}. However, such embedding-based strategies often introduce additional representation alignment overhead and may disrupt the original reasoning mechanisms of language models.
To address this issue, we adopt a simple yet effective format translating strategy that converts all implicit features into structured natural language descriptions. Through this translation process, implicit features are aligned with explicit contextual information in the same textual modality, enabling seamless integration into the original InstructTime prompt construction pipeline. This unified textual representation preserves the semantic content of implicit features while maintaining modality consistency, allowing language models to jointly model and reason over both explicit and implicit information without modifying their underlying architecture.
\subsubsection{Implicit Feature Fusion}
As illustrated in Figure~\ref{fig:Visualization}, the prompt template serves as a structured bridge between the translated textual representations and the language model. To effectively integrate heterogeneous features into a unified generative reasoning process, we design a structured prompt template that organizes explicit contextual cues and induced implicit features into a coherent textual form.
Specifically, the prompt template is composed of three main components: task-specific instructions, explicit contextual features, and implicit feature descriptions translated from statistical and visual toolkits. Each component is arranged in a predefined order with clear semantic boundaries, allowing the language model to distinguish their respective roles while jointly reasoning over all available information. This structured organization preserves the semantic integrity of implicit features and mitigates potential interference among heterogeneous inputs.
\revised{Regarding prompt sensitivity, the structured design ensures that each component plays a clearly defined role, such that reasonable variations in wording or ordering do not fundamentally alter the interpretation of the input. 
Regarding generalization, the overall prompt template remains consistent across different datasets, while only minimal adaptations (e.g., domain-specific descriptions in task instructions) are introduced when necessary. This ensures a unified modeling paradigm without relying on dataset-specific prompt engineering.
By adopting such a unified and structured prompt template, InstructTime++ enables language models to reason over both explicit and implicit information within a single textual context, without requiring any modification.}

\section{Experiments}
In this section, we first describe the experimental settings. We then evaluate the performance of InstructTime and InstructTime++ across multiple datasets and compare them with various baseline methods. In addition, we conduct further analytical experiments on InstructTime to examine the impact of its key components. Finally, we perform extended evaluations of InstructTime++ to assess its effectiveness.
Moreover, we present a case study with visualizations to further investigate the performance and properties of the proposed tokenizer.
\begin{table}[ht]
	\centering
	\caption{Statistics of the multiple datasets used in our experiments.}
	\resizebox{0.475\textwidth}{!}{%
		\begin{tabular}{c|ccccc}
			\toprule
			Datasets & \# Train Size & \# Test Size & \# Length & \# Channel & \# Label \\
			\midrule
			EEG   & 12,787 & 1,421 & 3,000 & 2     & 8 \\
			ECG   & 10,854 & 1,206 & 5,000 & 12    & 27 \\
			HAR   & 8,823 & 2,947 & 128   & 9     & 6 \\
			FD    & 10,912 & 2,728 & 5,120 & 1     & 3 \\
			RWC   & 10,934 & 1,962 & 4,000 & 1     & 2 \\
			\hline
			EP   & 9,200 & 2,300 & 178   & 1     & 2 \\
			SAD    & 5,940 & 1,980 & 93    & 13    & 10 \\
			\bottomrule
		\end{tabular}%
	}
	\label{tab:datasets}%
\end{table}%
\subsection{Experimental Settings}
\subsubsection{Datasets}

We conduct comprehensive evaluations on multiple widely used time series classification datasets, including electroencephalogram (EEG)~\cite{andrzejak2001indications}, electrocardiogram (ECG)~\cite{anguita2013public}, and human activity recognition (HAR)~\cite{cheng2025cross}. In addition, the fault detection (FD), right whale calls (RWC), epilepsy (EP), and spoken arabic digits (SAD) datasets are drawn from the UEA multivariate time series classification archive~\cite{bagnall2018uea}. Among them, the EP and SAD datasets are specifically used to evaluate the cross-domain transferability of InstructTime.
The key statistics of these datasets, including the number of instances, sequence length, and number of classes, are summarized in Table~\ref{tab:datasets}.
Below, we briefly describe each dataset.
The EEG dataset is based on the sleep-EDF database and contains full-night polysomnography recordings annotated with multiple sleep stages, capturing complex neurological dynamics.
The ECG dataset is derived from the physioNet challenge 2020 and focuses on multi-label classification of cardiac abnormalities across heterogeneous cohorts.
The HAR dataset consists of smartphone-based inertial sensor recordings from 30 participants performing six daily activities, serving as a standard benchmark for human activity recognition.
The FD dataset targets rolling bearing fault diagnosis in electromechanical drive systems, with labels indicating normal operation and different damage types.
The RWC dataset comprises real-world ocean acoustic recordings for binary detection of right whale vocalizations amid complex background noise.
The EP dataset consists of single-channel EEG time series collected from 500 subjects, segmented into 1-second samples and reformulated as a binary classification task.
The SAD dataset consists of 8,800 multivariate time series, each encoded as a 13-dimensional acoustic feature sequence obtained from spoken Arabic digit recordings of 88 native speakers.

\subsubsection{Baselines}
To evaluate the effectiveness of the proposed framework, we compare it with a set of representative baselines covering different modeling paradigms.
(1) Self-attention-based models include Transformer, Patch Transformer~\cite{zerveas2021transformer}, and FormerTime~\cite{cheng2023formertime}. These methods adapt the Transformer architecture to time series modeling by leveraging self-attention mechanisms to capture long-range temporal dependencies, with Patch Transformer improving computational efficiency through patch-wise segmentation and FormerTime further incorporating hierarchical multi-scale modeling strategies.
(2) Convolutional-based approaches consist of MCDCNN~\cite{zheng2014time}, TCN~\cite{oord2016wavenet}, and MiniROCKET~\cite{dempster2021minirocket}. These methods rely on convolutional operations to extract discriminative temporal patterns, where TCN employs dilated convolutions to model long-term dependencies efficiently and MiniROCKET transforms time series into compact feature representations for fast and accurate classification.
(3) Self-supervised learning models include TimeMAE~\cite{cheng2026timemae} and TS-TCC~\cite{eldele2021time}, which learn generic time series representations without label supervision through masked reconstruction and contrastive learning objectives, respectively.
(4) GPT-based model refers to GPT-As-Classifier, a novel adaptation of GPT2~\cite{radford2019language} in which the final layer is modified for time series classification, leveraging the representational capacity of large language models for sequential data.
For a fair comparison, Patch Transformer and GPT-As-Classifier employ identical input formats. In addition, the self-supervised models are instantiated with Transformer-based encoders, with specific adjustments to remove factors that could introduce unfair advantages.
We evaluate the proposed method under two settings: InstructTime-Universal, which excludes domain-specific fine-tuning, and InstructTime-Adapt, which incorporates both cross-domain and domain-specific fine-tuning.
These settings respectively reflect a general-purpose foundation model and a domain-adapted variant. Performance is assessed using Accuracy and F1 score.
\subsubsection{Implementation Details.}

All models are implemented using Python 3.11 and PyTorch 2.1.2, and experiments are conducted on Nvidia GeForce RTX 4090 GPUs. For baseline methods that do not involve PLMs, we adopt consistent hyper-parameter configurations to ensure fair comparison, including an embedding size of 64, a learning rate of 0.001, a batch size of 64, and the Adam optimizer with a weight decay of $1\times10^{-5}$.
For models that incorporate PLMs, the training strategy is adjusted to reflect their pre-trained initialization and larger parameter scale. We retain the default architecture of the PLMs, including the number of layers and embedding dimensions. During representation alignment pre-training, the learning rate is set to $5\times10^{-5}$ and reduced to $1\times10^{-5}$ for downstream tasks, with the batch size reduced to 16 to accommodate higher memory consumption. A warm-up strategy with a warm-up ratio of 0.05 and a cosine annealing scheduler is employed.
For InstructTime++, which is built upon the LLM Qwen3-0.6B, we further adjust the training configuration and conduct distributed training across two RTX 4090 GPUs using the NCCL backend. We adopt the AdamW optimizer with a weight decay of 0.01 and apply gradient clipping with a maximum norm of 1.0. The learning rate is set to $5\times10^{-4}$. Due to the extended encoder maximum length of 2048 tokens, the batch size is reduced to 4 per GPU. The model is trained for up to 30 epochs with early stopping (patience of 10 epochs) based on validation accuracy, while maintaining the same warm-up and cosine annealing scheduling strategy.

\begin{table*}[htbp]
	\centering
	\caption{Comparative classification results of all methods across multiple datasets. The best results are highlighted in \textbf{bold}, and the second-best results are \underline{underlined}.}
	\resizebox{1\textwidth}{!}{%
	\begin{tabular}{c|cccccccccc}
		\toprule
		\multirow{2}[3]{*}{Compared Models} & \multicolumn{2}{c}{EEG} & \multicolumn{2}{c}{ECG} & \multicolumn{2}{c}{HAR} & \multicolumn{2}{c}{FD} & \multicolumn{2}{c}{RWC} \\
		\cmidrule{2-11}
		& Accuracy & F1 Score & Accuracy & F1 Score & Accuracy & F1 Score & Accuracy & F1 Score & Accuracy & F1 Score \\
		\midrule
		Transformer 
        & 0.7940 & 0.5178 & 0.1821 & 0.3691 & 0.9148 & 0.9149 & 0.9451 & 0.9564 & 0.7158 & 0.7152 \\
		Patch Transformer 
        & 0.8076 & 0.5460 & 0.2465 & 0.3883 & 0.8704 & 0.8683 & 0.9390 & 0.9471 & 0.7552 & 0.7545 \\
		FormerTime 
        & \underline{0.8356} & \underline{0.5828} & 0.3712 & 0.5233 & 0.9199 & 0.9198 & 0.9732 & 0.9853 & \underline{0.7803} & \underline{0.7796} \\
		\midrule
		MCDCNN 
        & 0.8102 & 0.5395 & 0.0929 & 0.1735 & 0.8873 & 0.8862 & 0.9396 & 0.9545 & 0.7762 & 0.7759 \\
		TCN   
        & 0.7525 & 0.3927 & 0.1014 & 0.1654 & 0.9002 & 0.8997 & 0.7962 & 0.7248 & 0.7113 & 0.7109 \\
		MiniROCKET 
        & 0.8318 & 0.5638 & 0.2689 & 0.3900 & 0.9173 & 0.9153 & 0.9412 & 0.9569 & 0.7569 & 0.7556 \\
		\midrule
		TimeMAE 
        & 0.8248 & 0.5865 & 0.2546 & 0.3834 & \underline{0.9294} & \underline{0.9284} & 0.9878 & 0.9904 & 0.7690 & 0.7664 \\
		TS-TCC 
        & 0.7291 & 0.4347 & 0.1778 & 0.3780 & 0.8832 & 0.8815 & 0.9296 & 0.9363 & 0.6979 & 0.6931 \\
		\midrule
		GPT-As-Classifier 
        & 0.7689 & 0.4929 & 0.2253 & 0.3557 & 0.8973 & 0.8963 & 0.9489 & 0.9598 & 0.7554 & 0.7553 \\
		\midrule
		InstructTime-Universal 
        & 0.8067 & 0.5007 & 0.3402 & 0.4820 & 0.8990 & 0.8944 & 0.9619 & 0.9656 & 0.7307 & 0.7299 \\
		InstructTime-Adapt 
        & \underline{0.8452} & \underline{0.6240} & \underline{0.4121} & \underline{0.5547} & \textbf{0.9298} & \textbf{0.9307} & \underline{0.9901} & \underline{0.9917} & 0.7599 & 0.7578 \\
        \midrule
        InstructTime++ Universal 
        & 0.8311 & 0.5680 & \underline{0.4245} & \underline{0.5872} & 0.9036 & 0.9055 & 0.9831 & 0.9855 & 0.7497 & 0.7686 \\
        InstructTime++ Adapt 
        & \textbf{0.8776} & \textbf{0.6735} & \textbf{0.4693} & \textbf{0.6489} & 0.9257 & 0.9284 & \textbf{0.9934} & \textbf{0.9952} & \textbf{0.8048} & \textbf{0.8041} \\
		\bottomrule
	\end{tabular}%
	}
	\label{tab:main}
\end{table*}

\subsection{Classification Results Comparison}
Table~\ref{tab:main} comprehensively compares classification performance across diverse datasets. Overall, InstructTime++ Adapt achieves the best performance in most settings, while InstructTime-Adapt and InstructTime++ Universal consistently rank among the top performers. These results further confirm that instruction-driven generative paradigms provide an effective inductive bias for time series classification, particularly by enabling semantic alignment between numerical sequences and linguistic representations across diverse temporal domains.
Notably, self-supervised baselines such as TimeMAE demonstrate competitive performance on specific datasets (e.g., HAR), indicating that no single model universally dominates all scenarios and that different temporal structures favor different representation learning biases~\cite{cheng2023formertime}. In particular, datasets with relatively regular and repetitive temporal patterns can be effectively modeled through masked reconstruction objectives. However, the performance of such methods degrades on datasets characterized by higher semantic ambiguity and more subtle temporal dynamics, such as EEG and ECG, where discriminative cues are weak and class boundaries are semantically correlated. In these cases, purely reconstruction-driven or discriminative learning paradigms struggle to capture meaningful class semantics.
In contrast, InstructTime++ maintains consistently strong performance across all datasets. This robustness can be primarily attributed to its explicit modeling of implicit temporal and structural features, which complements explicit contextual cues within a unified generative framework. By translating both explicit and implicit information into a shared textual space and performing classification through semantic-aware label generation, InstructTime++ effectively reduces the representation gap between numerical sequences and linguistic semantics, leading to reliable and state-of-the-art performance across diverse classification settings.

\subsection{Experimental Analysis of InstructTime}
\subsubsection{Effect of Cross-domain Transferring}
\paragraph{Analysis of Cross-domain Pre-training}

As shown in Table ~\ref{tab:cross_domain}, cross-domain autoregressive pre-training consistently improves the performance of both InstructTime-Universal and InstructTime-Adapt across all datasets. The gains are particularly evident on EEG and ECG, where complex temporal dynamics and higher semantic ambiguity make domain-invariant representation learning more critical. Even on datasets with relatively regular temporal patterns, cross-domain pre-training yields stable improvements, indicating enhanced generalization rather than domain-specific overfitting. These results confirm that cross-domain pre-training effectively strengthens semantic alignment and improves generalization across heterogeneous time series domains. 
\begin{table*}[h]
	\centering
	\caption{Effect of cross-domain auto-regressive pre-training on InstructTime across multiple datasets.}
    
	\resizebox{1\textwidth}{!}{%
		\begin{tabular}{cc|cccccccccc}
			\toprule
			\multicolumn{1}{c}{\multirow{2}[2]{*}{\begin{tabular}{@{}c@{}}Models\end{tabular}}} & \multirow{2}[2]{*}{Settings} & \multicolumn{2}{c}{EEG} & \multicolumn{2}{c}{ECG} & \multicolumn{2}{c}{HAR} & \multicolumn{2}{c}{FD} & \multicolumn{2}{c}{RWC} \\
			\cmidrule{3-12}                            &       & Accuracy & F1 Score & Accuracy & F1 Score & Accuracy & F1 Score & Accuracy & F1 Score & Accuracy & F1 Score \\
			\midrule
			\multirow{2}[2]{*}{InstructTime-Universal} & w/o Cross-domain & 0.7854  & 0.4854  & 0.2554  & 0.3751  & 0.8341  & 0.8296  & 0.9092  & 0.9203  & 0.7270  & 0.7268  \\
			& w/ Cross-domain & \textbf{0.8067 } & \textbf{0.5007 } & \textbf{0.3402 } & \textbf{0.4820 } & \textbf{0.8990 } & \textbf{0.8944 } & \textbf{0.9619 } & \textbf{0.9656 } & \textbf{0.7307 } & \textbf{0.7299 } \\
			\midrule
			\multirow{2}[2]{*}{InstructTime-Adapt} & w/o Cross-domain & 0.8283  & 0.5751  & 0.3740  & 0.5209  & 0.9019  & 0.9028  & 0.9813  & 0.9830  & 0.7482  & 0.7478  \\
			& w/ Cross-domain & \textbf{0.8452 } & \textbf{0.6240 } & \textbf{0.4121 } & \textbf{0.5547 } & \textbf{0.9298 } & \textbf{0.9307 } & \textbf{0.9901 } & \textbf{0.9917 } & \textbf{0.7599 } & \textbf{0.7578 } \\
			\bottomrule
		\end{tabular}%
	}
	\label{tab:cross_domain}%
\end{table*}%
\paragraph{Analysis under Non-overlapping Domains}
Following the cross-domain pre-training results in Table~\ref{tab:cross_domain}, we further evaluate the effectiveness of pre-training under a more challenging non-overlapping domain setting. As shown in Table~\ref{tab:non-overlap}, pre-training continues to yield substantial performance improvements on both EP and SAD datasets, even though these domains are entirely unseen during pre-training. The pronounced gains, particularly in F1 score, indicate that the learned representations capture transferable semantic and temporal patterns rather than domain-specific characteristics. These results further corroborate that cross-domain pre-training enables InstructTime to generalize effectively to unseen domains, reinforcing its applicability in realistic and practically relevant transfer learning scenarios.
\begin{table}[h]
	\centering
    \small
	\caption{Effect of pre-training on transfer learning across two external domains.}
	\begin{tabular}{c|cccc}
		\toprule
		\multirow{2}[2]{*}{Model Variants} & \multicolumn{2}{c}{EP} & \multicolumn{2}{c}{SAD} \\
		\cmidrule{2-5}  & Accuracy & F1 Score & Accuracy & F1 Score \\
		\midrule
		w/o Pre-training & 0.8896  & 0.5558  & 0.8045  & 0.7199  \\
		w/ Pre-training & \textbf{0.9596 } & \textbf{0.9355 } & \textbf{0.9237 } & \textbf{0.9234 } \\
		\bottomrule
	\end{tabular}%
	\label{tab:non-overlap}%
\end{table}%
\begin{table*}[h]
	\centering
    \caption{Effect of autoregressive pre-training on InstructTime across multiple datasets.}
    \resizebox{\textwidth}{!}{%
	\begin{tabular}{c|cccccccccc}
    
		\toprule
		\multirow{2}[4]{*}{Model Variants} & \multicolumn{2}{c}{EEG} & \multicolumn{2}{c}{ECG} & \multicolumn{2}{c}{HAR} & \multicolumn{2}{c}{FD} & \multicolumn{2}{c}{RWC} \\
		\cmidrule{2-11}          & Accuracy & F1 Score & Accuracy & F1 Score & Accuracy & F1 Score & Accuracy & F1 Score & Accuracy & F1 Score \\
		\midrule
		w/o Pre-training & 0.7854  & 0.4854  & 0.2554  & 0.3751  & 0.8341  & 0.8296  & 0.9092  & 0.9203  & 0.7270  & 0.7268  \\
		w/ Pre-training & \textbf{0.8452 } & \textbf{0.6240 } & \textbf{0.4121 } & \textbf{0.5547 } & \textbf{0.9298 } & \textbf{0.9307 } & \textbf{0.9901 } & \textbf{0.9917 } & \textbf{0.7599 } & \textbf{0.7578 } \\
		\bottomrule
	\end{tabular}%
    }
	\label{tab:ar}%
\end{table*}%
\subsubsection{Effect of Auto-regressive Pre-training}
As shown in Table~\ref{tab:ar}, auto-regressive pre-training leads to consistent performance improvements for InstructTime across all evaluated datasets. Notably, substantial gains are observed on EEG and ECG, where modeling long-range temporal dependencies and complex temporal dynamics is particularly important, indicating that autoregressive objectives effectively enhance the quality of learned temporal representations. Even on datasets with relatively regular temporal patterns, such as HAR and FD, autoregressive pre-training yields stable improvements in both Accuracy and F1 score, suggesting that its benefits are not limited to highly complex domains. This observation implies that autoregressive pre-training helps capture general temporal structures that are broadly useful across different data characteristics. Moreover, by encouraging the model to predict future temporal patterns in a sequential manner, autoregressive pre-training provides a stronger inductive bias for temporal continuity and dependency modeling. Overall, these results demonstrate that autoregressive pre-training plays a critical role in strengthening temporal modeling capability and improving classification performance across different datasets.
\begin{table*}[h]
	\centering
    \caption{Impact of instruction text on modality alignment under auto-regressive training.}
	\resizebox{1\textwidth}{!}{%
		\begin{tabular}{c|cccccccccc}
			\toprule
			\multirow{2}[2]{*}{Model Variants} & \multicolumn{2}{c}{EEG} & \multicolumn{2}{c}{ECG} & \multicolumn{2}{c}{HAR} & \multicolumn{2}{c}{FD} & \multicolumn{2}{c}{RWC} \\
			\cmidrule{2-11}                            & Accuracy & F1 Score & Accuracy & F1 Score & Accuracy & F1 Score & Accuracy & F1 Score & Accuracy & F1 Score \\
			\midrule
			w/o Text & 0.7748  & 0.4641  & 0.1675  & 0.1779  & 0.8697  & 0.8676  & 0.8937  & 0.8984  & 0.7492  & 0.7489  \\
			w/  Text & \textbf{0.8452 } & \textbf{0.6240 } & \textbf{0.4121 } & \textbf{0.5547 } & \textbf{0.9298 } & \textbf{0.9307 } & \textbf{0.9901 } & \textbf{0.9917 } & \textbf{0.7599 } & \textbf{0.7578 } \\
			\bottomrule
		\end{tabular}%
	}
	\label{tab:text}%
\end{table*}%
\subsubsection{Analysis of Instruction Text in Modality Alignment}

\begin{figure}[h]
	\centering
	\includegraphics[width=0.55\textwidth]{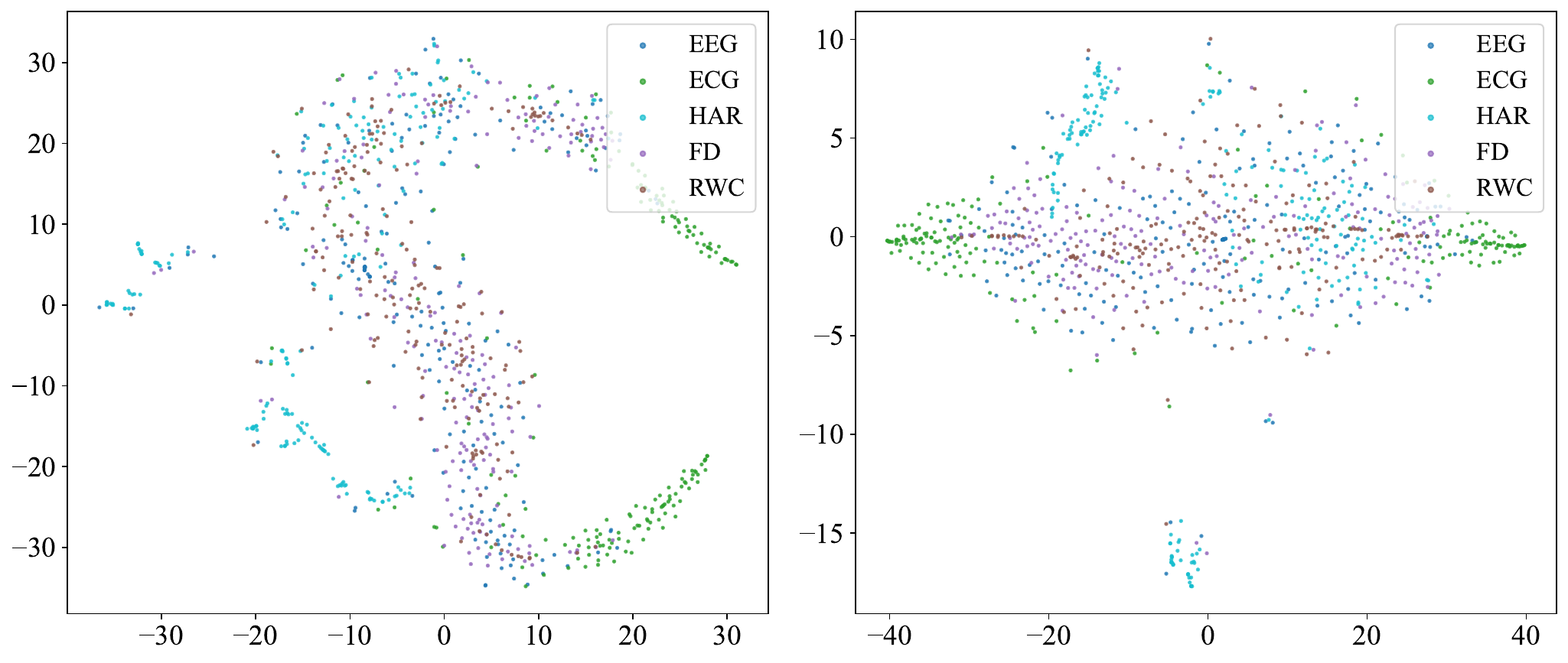}
	\caption{Visualization of t-SNE embeddings for instances sampled from multiple datasets.}
	\label{fig:tsne_five}
\end{figure}

As shown in Table~\ref{tab:text}, preserving instruction text during autoregressive training leads to consistent and substantial performance improvements across all evaluated datasets. When textual guidance is removed, performance degrades markedly, indicating that instruction text plays a crucial role in aligning numerical time series representations with the underlying semantic task descriptions. Incorporating instruction text yields stable gains in both Accuracy and F1 score, suggesting improved semantic consistency rather than mere optimization effects.
Figure~\ref{fig:tsne_five} provides complementary qualitative evidence. Without textual prompts, embeddings from different domains exhibit considerable overlap and dispersion, reflecting weak semantic alignment in the learned representation space. In contrast, preserving instruction text results in more compact and structured embeddings with clearer domain-wise organization. This qualitative observation aligns well with the quantitative improvements in Table~\ref{tab:text}, collectively demonstrating that instruction text effectively facilitates modality alignment and enhances the semantic coherence of learned time series representations.

\subsubsection{Analysis of Multi-class and Multi-label Classification}
\begin{figure}[h]
	\centering
	\includegraphics[width=1\textwidth]{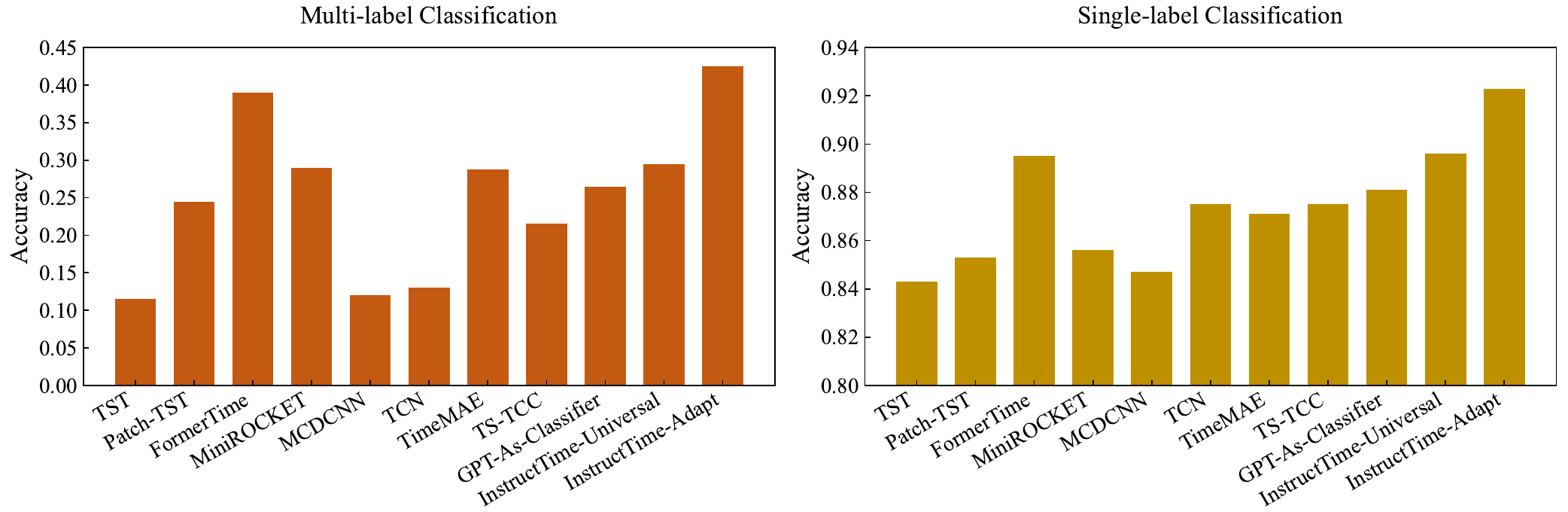}
	\caption{Performance comparison of multi-label and single-label ECG classification.}
	\label{fig:multi-label}
\end{figure}

Figure~\ref{fig:multi-label} illustrates the performance of different models on complex multi-label and single-label classification tasks evaluated on the ECG dataset. Most approaches achieve relatively strong performance in the multi-label classification, which is expected since multi-label classification is more challenging. Notably, InstructTime-Adapt consistently outperforms all compared methods by achieving the highest accuracy across both settings. Moreover, the performance gains over baseline methods are more pronounced in the multi-label scenario, indicating the effectiveness of InstructTime-Adapt in handling complex label structures. We attribute this improvement to the inductive biases introduced by multi-label textual descriptions, which are effectively captured through the proposed generative framework.

\subsubsection{Analysis of Data Efficiency}

Table~\ref{tab:few-shot} reports the results of data-efficient evaluation under the few-shot setting~\cite{brown2020language}, where the proportion of the training set is progressively reduced to $70\%$, $40\%$, and $10\%$. Two model variants are evaluated: one trained from scratch and one initialized with pre-training. This setting allows us to examine the model’s ability to maintain accurate classification performance under extremely limited training data, which is particularly relevant in data-sparse scenarios. As the proportion of training data increases from $10\%$ to $70\%$, a consistent performance improvement is observed across all datasets, indicating that larger training sets generally contribute to better model performance. However, the degree of improvement varies across datasets. 
For example, the HAR dataset shows a larger performance gain than the ECG dataset, likely because pre-training provides a stronger initialization for data with clearer motion patterns.
\begin{table*}[htbp]
	\centering
	\caption{
	Experimental results of our InstructTime-Universal tuned with different levels of training set.
	}
	\resizebox{1\textwidth}{!}{%
		\begin{tabular}{cc|cccccccccc}
			\toprule
			\multirow{2}{*}{\begin{tabular}{@{}c@{}}Model\\Variants\end{tabular}} 
			& \multirow{2}{*}{Proportion} 
			& \multicolumn{2}{c}{EEG} 
			& \multicolumn{2}{c}{ECG} 
			& \multicolumn{2}{c}{HAR} 
			& \multicolumn{2}{c}{FD} 
			& \multicolumn{2}{c}{RWC} \\
			\cmidrule{3-12}
			& & Accuracy & F1 Score & Accuracy & F1 Score & Accuracy & F1 Score & Accuracy & F1 Score & Accuracy & F1 Score \\
			\midrule
			
			\multirow{3}{*}{\begin{tabular}{@{}c@{}}w/o\\Pre-training\end{tabular}}
			& 10\%  
			& 0.6151 & 0.3061 
			& 0.1467 & 0.1538 
			& 0.6512 & 0.6317 
			& 0.6979 & 0.5319 
			& 0.5367 & 0.3322 \\
			
			& 40\%  
			& 0.6685 & 0.3163 
			& 0.1318 & 0.2321 
			& 0.7336 & 0.7160 
			& 0.7309 & 0.5720 
			& 0.5107 & 0.3451 \\
			
			& 70\%  
			& \underline{0.7431} & \underline{0.4219} 
			& 0.1882 & 0.2664 
			& 0.7512 & 0.7346 
			& 0.7639 & 0.5936 
			& 0.5724 & 0.4146 \\
			
			\midrule
			
			\multirow{3}{*}{\begin{tabular}{@{}c@{}}w/\\Pre-training\end{tabular}}
			& 10\%  
			& 0.6974 & 0.4070 
			& 0.3781 & 0.5185 
			& 0.7774 & 0.6627 
			& 0.8244 & 0.6251 
			& 0.7044 & 0.6825 \\
			
			& 40\%  
			& 0.7368 & 0.4666 
			& \underline{0.3922} & \underline{0.5421} 
			& \underline{0.8357} & \underline{0.7140} 
			& \underline{0.8831} & \underline{0.6780} 
			& \underline{0.7253} & \underline{0.7008} \\
			
			& 70\%  
			& \textbf{0.7607} & \textbf{0.4728} 
			& \textbf{0.3972} & \textbf{0.5491} 
			& \textbf{0.8571} & \textbf{0.7337} 
			& \textbf{0.8842} & \textbf{0.9063} 
			& \textbf{0.7080} & \textbf{0.7088} \\
			
			\bottomrule
		\end{tabular}%
	}
	\label{tab:few-shot}
\end{table*}

\begin{table*}[htbp]
	\centering
	\caption{Performance of InstructTime-Universal under different projection layer configurations.}
	\resizebox{0.975\textwidth}{!}{%
		\begin{tabular}{cc|cccccccccc}
			\toprule
			\multirow{2}[2]{*}{\begin{tabular}{@{}c@{}}Alignment\\Projector\end{tabular}} & \multirow{2}[2]{*}{Hidden Size} & \multicolumn{2}{c}{EEG} & \multicolumn{2}{c}{ECG} & \multicolumn{2}{c}{HAR} & \multicolumn{2}{c}{FD} & \multicolumn{2}{c}{RWC} \\
			\cmidrule{3-12}                            &       & Accuracy & F1 Score & Accuracy & F1 Score & Accuracy & F1 Score & Accuracy & F1 Score & Accuracy & F1 Score \\
			\midrule
			Linear & 64, 768 & 0.7776  & 0.4287  & 0.2699  & 0.3410  & 0.8614  & 0.8532  & 0.9249  & 0.9310  & 0.7118  & 0.7082  \\
			\midrule
			\multirow{2}[2]{*}{MLP} & 128, 512, 768 & 0.7935  & 0.4849  & 0.3060  & 0.4451  & 0.8885  & 0.8828  & \textbf{0.9685 } & \textbf{0.9714 } & 0.7227  & 0.7121  \\
			& 64, 128, 256, 512, 768 & \textbf{0.8067 } & \textbf{0.5007 } & \textbf{0.3402 } & \textbf{0.4820 } & \textbf{0.8990 } & \textbf{0.8944 } & 0.9619  & 0.9656  & \textbf{0.7307 } & \textbf{0.7299 } \\
			\bottomrule
		\end{tabular}%
	}
	\label{tab:projector}%
\end{table*}%

\subsubsection{Analysis of Alignment Projector}
Table~\ref{tab:projector} displays the performance of InstructTime model variants with different projection layer configurations, specifically comparing a linear alignment projector against an MLP (multi-layer perceptron) with various hidden sizes. Across datasets such as EEG, ECG, HAR, FD, and RWC, it is evident that the MLP projector with a more complex hidden size configuration significantly enhances both accuracy and F1 scores. In summary, the experimental results underscore the importance of the projection layer's configuration in the InstructTime model, with a more nuanced MLP projector leading to significant performance improvements across multiple datasets. For instance, the MLP projector with hidden sizes of 64, 128, 256, 512, and 768 achieves the highest performance metrics. This trend is consistent across other datasets, with the MLP projector outperforming the Linear one, highlighting the benefit of employing a more complex projection layer in capturing and processing the data's underlying patterns more effectively.

\begin{table*}[htbp]
	\centering
	\caption{Effect of discretization token number in InstructTime-Universal. }
	\resizebox{1\textwidth}{!}{%
		\begin{tabular}{c|ccccccccccccccc}
			\toprule
			\multirow{2}[2]{*}{Num. Token} 
			& \multicolumn{3}{c}{EEG} 
			& \multicolumn{3}{c}{ECG} 
			& \multicolumn{3}{c}{HAR} 
			& \multicolumn{3}{c}{FD} 
			& \multicolumn{3}{c}{RWC} \\
			\cmidrule{2-16}
			& MSE & Accuracy & F1 Score 
			& MSE & Accuracy & F1 Score 
			& MSE & Accuracy & F1 Score 
			& MSE & Accuracy & F1 Score 
			& MSE & Accuracy & F1 Score \\
			\midrule
			128  
			& 0.0684 & 0.7847 & 0.4586
			& 0.0732 & \textbf{0.2475} & \textbf{0.3569}
			& 0.0579 & 0.7766 & 0.7610
			& 0.0732 & 0.8455 & 0.8369
			& 0.0862 & 0.7013 & 0.6994 \\

			256  
			& 0.0595 & \underline{0.7931} & \textbf{0.5122}
			& 0.0769 & \underline{0.2442} & \underline{0.3394}
			& 0.0501 & \textbf{0.8695} & \textbf{0.8683}
			& 0.0769 & 0.8462 & 0.8205
			& \underline{0.1998} & \underline{0.7194} & \underline{0.7183} \\

			384  
			& \textbf{0.0571} & 0.7882 & \underline{0.4911}
			& 0.0766 & 0.2206 & 0.3032
			& 0.0459 & \underline{0.8643} & \underline{0.8000}
			& 0.0766 & \underline{0.8613} & \underline{0.8883}
			& 0.2024 & \textbf{0.7314} & \textbf{0.7311} \\

			512  
			& 0.0579 & 0.7822 & 0.4739
			& \underline{0.0711} & 0.1870 & 0.2980
			& 0.0471 & 0.8439 & 0.7812
			& \underline{0.0711} & \textbf{0.9245} & \textbf{0.9301}
			& \textbf{0.1993} & 0.7148 & 0.6003 \\

			768  
			& \underline{0.0575} & \textbf{0.7945} & \underline{0.5000}
			& \textbf{0.0684} & 0.2388 & \underline{0.3542}
			& \textbf{0.0433} & 0.6861 & 0.6603
			& \textbf{0.0684} & 0.8880 & 0.8915
			& 0.2117 & 0.6934 & 0.6874 \\
			\bottomrule
		\end{tabular}%
	}
	\label{tab:token_number}
\end{table*}

\subsubsection{Analysis of Codebook Size}
Table~\autoref{tab:token_number} reports the effect of the number of discretization tokens used in the tokenization stage of InstructTime, evaluated using MSE, Accuracy, and F1 Score. The results reveal a clear and consistent pattern: neither the smallest nor the largest token numbers yield optimal performance across datasets. Instead, intermediate token sizes generally achieve better results, indicating a favorable trade-off between representation granularity and modeling capacity. Using too few tokens limits the expressiveness of discretized representations, while excessively large token sets introduce redundancy and increase optimization difficulty. These observations suggest the existence of a practical “sweet spot” in token number selection, which effectively preserves essential information while maintaining model stability and efficiency.

\subsubsection{Analysis of Token Distribution}
\begin{figure*}[t]
	\centering
	\includegraphics[width=1.0\textwidth]{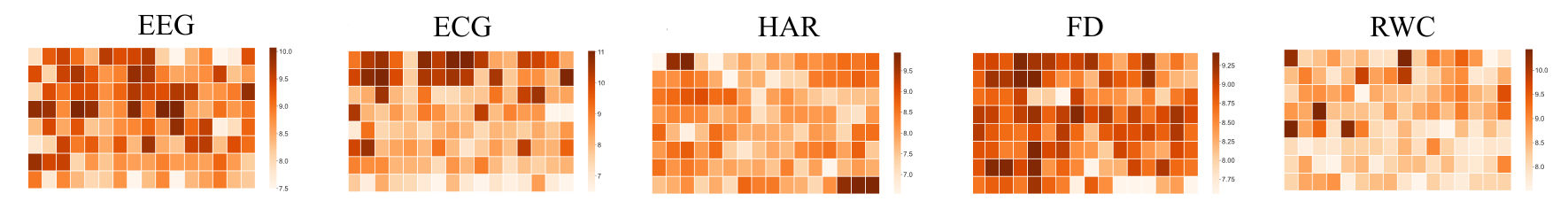}
	\caption{Statistics of token usage frequency across multiple datasets. Each heatmap corresponds to one dataset (EEG, ECG, HAR, FD, and RWC), where color intensity indicates the frequency of token occurrence.}
	\label{fig:token}
\end{figure*}

Figure~\ref{fig:token} presents heatmap visualizations of token usage frequency across five different domains. The results show that token usage distributions vary notably across datasets. Datasets such as HAR and RWC exhibit relatively uniform token usage, as reflected by more homogeneous color patterns, suggesting a balanced distribution of feature states captured by the tokens. In contrast, datasets including EEG, ECG, and FD display higher variance in token usage, with certain tokens appearing more frequently, indicated by concentrated regions of darker colors. This pattern implies that specific states or features are more dominant in these domains, or that the underlying data contain a higher degree of redundancy. Overall, the token frequency heatmaps indicate that different domains exhibit distinct levels of feature diversity and redundancy, highlighting the domain-dependent nature of token utilization in the tokenization process and its impact on representation learning.

\subsection{Extended Experimental Analysis of InstructTime++}

\subsubsection{Effect of Implicit Feature Modeling}
Table ~\ref{tab:multi-features} presents an ablation study examining the contributions of different implicit feature components in InstructTime++. Removing all implicit features leads to a clear performance degradation across all datasets and metrics, demonstrating that implicit features provide essential complementary information beyond explicit inputs. When ablating individual components, both visual text features and statistical text features consistently contribute to performance improvements, while their impact varies across domains. Specifically, removing visual text features results in more noticeable performance drops on datasets such as EEG and HAR, indicating that perception-driven semantic cues are particularly beneficial for capturing complex temporal patterns. In contrast, removing statistical text features leads to larger degradations on ECG and RWC, suggesting that rule-driven statistical descriptors play a more critical role in domains with strong numerical regularities. Notably, the full InstructTime++-Adapt model consistently achieves the best performance across all datasets, confirming that visual and statistical implicit features are complementary rather than redundant. These results validate the design choice of jointly modeling multiple types of implicit features to enhance reliability and generalization across heterogeneous time series domains.
\begin{table*}[htbp]
	\centering
	\caption{Effect of different implicit feature components in InstructTime++.}
	\resizebox{1\textwidth}{!}{%
		\begin{tabular}{c|cccccccccc}
			\toprule
			\multirow{2}[2]{*}{Model Variants} & \multicolumn{2}{c}{EEG} & \multicolumn{2}{c}{ECG} & \multicolumn{2}{c}{HAR} & \multicolumn{2}{c}{FD} & \multicolumn{2}{c}{RWC} \\
			\cmidrule{2-11}                           
            & Accuracy & F1 Score & Accuracy & F1 Score & Accuracy & F1 Score & Accuracy & F1 Score & Accuracy & F1 Score \\
            \midrule
            w/o Implicit Features 
            & 0.8283 & 0.5751 
            & 0.3740 & 0.5209 
            & 0.9019 & 0.9028 
            & 0.9813 & 0.9830 
            & 0.7482 & 0.7478 \\

            w/o Visual Features 
            & \underline{0.8600} & \underline{0.6409} 
            & 0.3964 & \underline{0.5565} 
            & \underline{0.9226} & \underline{0.9249} 
            & 0.9835 & \underline{0.9858} 
            & 0.7640 & 0.7640 \\

            w/o Statistical Features  
            & 0.8571 & 0.5354 
            & \underline{0.4104} & \underline{0.5831} 
            & 0.9172 & 0.9190 
            & \underline{0.9883} & \underline{0.9914} 
            & \underline{0.8017} & \underline{0.8017} \\

            InstructTime++ Adapt 
            & \textbf{0.8776} & \textbf{0.6735} 
            & \textbf{0.4693} & \textbf{0.6489} 
            & \textbf{0.9257} & \textbf{0.9284} 
            & \textbf{0.9934} & \textbf{0.9952} 
            & \textbf{0.8048} & \textbf{0.8041} \\
			\bottomrule
		\end{tabular}%
	}
	\label{tab:multi-features}%
\end{table*}

\subsubsection{Effect of Backbone Scaling}
Table~\ref{tab:backbone} investigates the effect of backbone size on the performance of InstructTime++ across different datasets. Overall, the results do not exhibit a strictly monotonic performance improvement with increasing backbone size. Instead, the medium-sized backbone, Qwen3-0.6B, achieves the best or near-best performance on most datasets and metrics, particularly on EEG, HAR, and RWC, indicating that a moderate model capacity is sufficient to capture the semantic and temporal patterns required by these tasks. While larger backbones such as Qwen3-1.7B and Qwen3-4B show competitive results and yield improvements on certain metrics (e.g., ECG accuracy and FD F1 score), their gains are not consistent across all datasets. It is worth noting that this comparison is conducted under a supervised fine-tuning setting only, due to computational resource constraints. This suggests that simply scaling up the backbone does not uniformly translate into better downstream performance and may introduce redundancy or optimization challenges. These findings imply that InstructTime++ benefits from an appropriate balance between representational capacity and task complexity, and that a moderately sized backbone can offer a more favorable trade-off between performance and efficiency across diverse time series domains.
             

\begin{table*}[htbp]
	\centering
	\caption{Effect of backbone size on InstructTime++ performance.}
	\resizebox{1\textwidth}{!}{%
		\begin{tabular}{c|cccccccccc}
			\toprule
			\multirow{2}[2]{*}{Model Variants} & \multicolumn{2}{c}{EEG} & \multicolumn{2}{c}{ECG} & \multicolumn{2}{c}{HAR} & \multicolumn{2}{c}{FD} & \multicolumn{2}{c}{RWC} \\
			\cmidrule{2-11}                           & Accuracy & F1 Score & Accuracy & F1 Score & Accuracy & F1 Score & Accuracy & F1 Score & Accuracy & F1 Score \\
			\midrule
			GPT-2  
            & 0.8093 & 0.5647 
            & 0.2993 & 0.4558 
            & 0.9033 & 0.9026 
            & 0.9740 & 0.9804 
            & 0.7686 & 0.7686 \\
            Qwen3-0.6B 
            & \textbf{0.8600} & \textbf{0.6409} 
            & \underline{0.4138} & \textbf{0.5709} 
            & \textbf{0.9247} & \underline{0.9274} 
            & \textbf{0.9857} & 0.9739 
            & \textbf{0.8038} & \textbf{0.8029} \\
             
            Qwen3-1.7B 
            & \underline{0.8593} & 0.5788 
            & 0.4088 & 0.5428 
            & 0.9155 & 0.9177 
            & \underline{0.9839} & \underline{0.9839} 
            & \underline{0.8002} & \underline{0.8002} \\
            Qwen3-4B  
            & 0.8578 & \underline{0.5963} 
            & \textbf{0.4146} & \underline{0.5633} 
            & \underline{0.9190} & \textbf{0.9323} 
            & 0.9824 & \textbf{0.9985} 
            & 0.8000 & 0.8001 \\
			\bottomrule
		\end{tabular}%
	}
	\label{tab:backbone}%
\end{table*}%

\subsubsection{Effect of Patch Size.}
Table~\ref{tab:patch_size} reports the experimental results evaluating the impact of different patch sizes on the performance of the InstructTime-Universal method across multiple datasets, along with the reconstruction performance of the VQ networks measured by mean squared error (MSE). The results show that patch size exhibits a non-uniform impact across datasets and evaluation objectives. In particular, smaller patch sizes generally lead to better reconstruction performance, indicating their effectiveness in capturing fine-grained temporal details. In contrast, larger patch sizes tend to yield improved classification performance on datasets such as FD and RWC, suggesting that coarser patches are more suitable for aggregating discriminative features relevant to downstream classification tasks. These observations imply a trade-off between reconstruction fidelity and classification effectiveness, depending on the chosen patch size. Moreover, no single patch size consistently outperforms others across all datasets and metrics, highlighting the necessity of dataset-specific configuration. Overall, the results indicate that patch size is a critical hyper-parameter in InstructTime-Universal that requires careful tuning to balance reconstruction accuracy and classification performance for different types of time series data.
\begin{table*}[htbp]
	\centering
	\caption{Effect of patch size in the tokenization stage of InstructTime-Universal.}
	\resizebox{0.975\textwidth}{!}{%
		\begin{tabular}{c|ccccccccccccccc}
			\toprule
			\multirow{2}[2]{*}{Patch Size} & \multicolumn{3}{c}{EEG} & \multicolumn{3}{c}{ECG} & \multicolumn{3}{c}{HAR} & \multicolumn{3}{c}{FD} & \multicolumn{3}{c}{RWC} \\
			\cmidrule{2-16}                            & MSE   & Accuracy & F1 Score & MSE   & Accuracy & F1 Score & MSE   & Accuracy & F1 Score & MSE   & Accuracy & F1 Score & MSE   & Accuracy & F1 Score \\
			\midrule
			25,20,1,32,25 & \textbf{0.0534 } & 0.7889  & 0.4976  & \textbf{0.0227 } & 0.3735  & 0.2354  & \textbf{0.0501 } & \textbf{0.8695 } & \textbf{0.8683 } & \textbf{0.0597 } & 0.8702  & 0.8724  & \textbf{0.1941 } & 0.7113  & 0.7079  \\
			40,25,2,40,32 & 0.0595  & 0.7906  & 0.5085  & 0.0265  & 0.3569  & 0.2067  & 0.0557  & 0.6860  & 0.5933  & 0.0711  & \textbf{0.9205 } & \textbf{0.9314 } & 0.2024  & \textbf{0.7314 } & \textbf{0.7311 } \\
			50,30,4,64,40 & 0.0651  & \textbf{0.7931 } & \textbf{0.5122 } & 0.0296  & \textbf{0.3975 } & \textbf{0.2421 } & 0.0646  & 0.7486  & 0.6656  & 0.0851  & 0.8308  & 0.8516  & 0.2293  & 0.7021  & 0.7012  \\
			\bottomrule
		\end{tabular}%
	}
	\label{tab:patch_size}%
\end{table*}%

\subsubsection{Analysis of VQ Networks Reconstructive Quality} 
\begin{figure*}[h]
	\centering
	\includegraphics[width=1.0\textwidth]{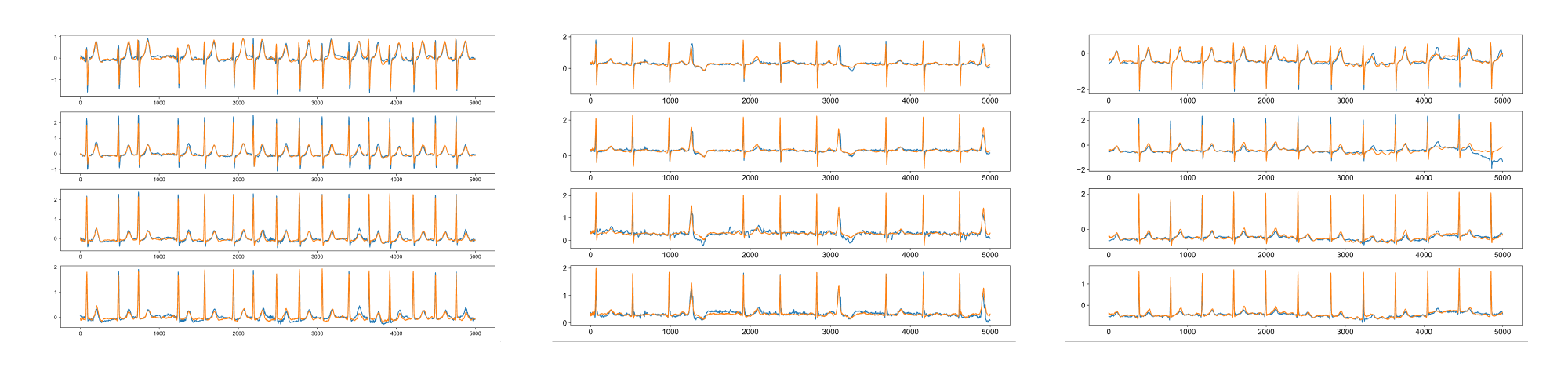}
	\caption{Visualizing the reconstruction of ECG signals in the vector quantized networks.}
	\label{fig:reconstruction}
\end{figure*}

Figure \ref{fig:reconstruction} offers a comparative illustration of the original and reconstructed ECG signals within the Vector Quantized (VQ) networks. Each subplot presents
two overlaid ECG traces: the original signal in blue and its reconstructed counterpart in orange. A visual inspection indicates that the VQ networks capture the fundamental morphology of the ECG time series effectively, with the P, QRS, and T waves being discernible in the reconstructed signal. However, there are noticeable deviations in certain segments where the reconstruction does not perfectly align with the original trace, particularly evident in the
amplitude and sharpness of the R wave peaks. These discrepancies might be attributed to the quantization error inherent to the VQ networks, which can lead to a loss of fine detail in the signal. Despite these imperfections, the overall time series patterns and intervals appear to be preserved. The consistency of the reconstructions across multiple cases suggests that the VQ networks have learned a generalized representation of the ECG data, which could
be valuable to use in the multimodal modeling problem.
\section{Related Work}
\label{sec:related_work}
In this section, we discuss two closely related research areas: time series classification and the application of language models to time series classification.
\subsection{Conventional Time Series Classification}
TSC has garnered significant attention from researchers in recent years~\cite{middlehurst2023bake,shifaz2020ts}. Through an extensive review of the literature, these contributions can be broadly categorized into four types of classical approaches.  Distance-based methods form the bedrock of traditional TSC, primarily leveraging similarity measures to classify time series. A typical example is the Dynamic Time Warping (DTW) algorithm, which, when combined with a nearest neighbor classifier (NN-DTW)~\cite{ding2008querying}, serves as a benchmark in the field~\cite{petitjean2014dynamic}. 
Shapelet-based~\cite{ye2009time} methods revolve around identifying sub-sequences (shapelets) that are predictive of a time series class. The learning shapelets~\cite{grabocka2014learning} algorithm is a prime example, providing an automated process for shapelet discovery directly from the data.
Dictionary-based techniques involve transforming time series into symbolic representations and then analyzing the frequency of these symbolic patterns. One of the seminal works in this area is the symbolic aggregate approximation (SAX)~\cite{lin2007experiencing}, which enables a bag-of-words type model for TSC. This approach is further extended by methods such as bag of patterns (BOP)~\cite{lin2012rotation} and symbolic aggregate approximation–vector space model (SAX-VSM)~\cite{senin2013sax}, which demonstrate effectiveness in capturing repetitive and local patterns. 
Feature-based approaches focus on extracting features from specific intervals of the time series. Time series forest~\cite{deng2013time} is a notable method in this category, employing random interval selection and summary statistics to capture local patterns.
Although effective in some scenarios, these methods excel in handling temporal distortions but can be computationally intensive for large datasets.
In contrast, Deep Learning-based methods gain increasing popularity due to their ability to efficiently learn complex features from raw time series~\cite{zhang2020tapnet,liu2024adaptive}.  CNNs~\cite{zheng2014time,ismail2020inceptiontime}, GNNs~\cite{han2021dynamic}, and self-attention based Transformer architectures~\cite{liu2021gated,cheng2023formertime} stand at the forefront of this wave, showcasing remarkable performance across diverse TSC tasks.  These deep learning methods benefit from the ability to model long-term dependencies and hierarchical feature representations without the need for manual feature engineering~\cite{christ2018time, chen2023adversarial}. Despite the profound nonlinear modeling capabilities of deep neural networks and their advantage of obviating the need for manual feature engineering, thereby facilitating the learning of more complex temporal features for effective classification—their major drawback lies in their voracious data appetite~\cite{wang2024neuralreconciler, wi2024continuous}. These models necessitate extensive labeled training sets, without which they are prone to overfitting~\cite{zhong2026logicgate}. 
\subsection{LM-based Time Series Classification}
In the realm of natural language processing (NLP), the advent of language models has marked a paradigm shift from traditional static word vector representations, such as Word2Vec~\cite{mikolov2013efficient}, to dynamic, context-aware embeddings. PLM leverages vast amounts of textual data to learn rich, nuanced language representations before being fine-tuned for specific tasks, embodying a more holistic approach to language understanding. The evolution of PLM can be broadly categorized into three architectural paradigms: decoder-only, encoder-only, and encoder-decoder models. Each of these paradigms brings its unique approach to language modeling and representation. (1) Decoder-only models, such as GPT~\cite{radford2018improving}, focus on generating text based on the preceding context. This design is particularly adept at tasks involving language generation, where the model predicts subsequent tokens given a sequence of prior tokens. GPT and its successors exemplify this approach, demonstrating remarkable proficiency in text completion, creative writing, and more. (2) Encoder-only models, epitomized by BERT~\cite{devlin2018bert}, specialize in understanding and interpreting the context of a given text fragment. By processing text in a bidirectional manner, these models excel at tasks requiring deep contextual understanding, such as sentiment analysis, named entity recognition, and question answering. (3) Encoder-decoder models, like BART~\cite{lewis2019bart}, combine the strengths of both encoders and decoders to handle a wide range of tasks from translation to summarization. These models are designed to encode the input text into an intermediate representation, which the decoder then uses to generate the output text, making them versatile tools for both understanding and generation tasks.
Building upon these PLM architectures, recent research advances toward LLMs by substantially scaling model parameters, training data, and optimization strategies. LLMs inherit the architectural foundations of PLMs while exhibiting emergent capabilities, such as in-context learning, multi-step reasoning, and instruction following. In the time series domain, recent LLM-based forecasting methods further investigate how contextual signals, symbolic representations, and memory mechanisms can support temporal reasoning~\cite{tao2026memcast}.

Unlike traditional PLMs that often rely on task-specific fine-tuning, LLMs demonstrate the ability to adapt to new tasks through natural language prompts, reducing the dependence on explicit supervised training.
To further enhance the controllability and usability of LLMs, several adaptation strategies are explored. Fine-tuning~\cite{devlin2018bert} slightly adjusts pre-trained parameters for specific objectives, while prompt engineering~\cite{brown2020language} reformulates tasks as textual instructions. Instruction tuning~\cite{yuan2025watermarking, ouyang2022training} aligns model behaviors with human intent by training on curated instruction–response pairs, and reinforcement learning from human feedback (RLHF) further refines generation quality and preference alignment through interactive feedback signals.

\section{Conclusion}

In this work, we examine the limitations of the dominant learning-to-classify paradigm in time series classification, particularly its inability to effectively incorporate contextual features and model semantic relationships among classes. To address these challenges, we propose InstructTime, a novel framework that reformulates time series classification as an instruction-driven learning-to-generate task. By fully exploiting the generative capability of language models, InstructTime enables generative time series classification through discrete prompt construction, representation alignment pre-training, and supervised fine-tuning.

Building upon this foundation, we further extend InstructTime to InstructTime++, which introduces an improved
multimodal generative framework, enhanced with implicit feature modeling. Specifically, we employ toolkits to extract statistical and visual features from raw time series or contextual features, and unify them into a natural language description. This design maintains high compatibility with the  original framework while effectively compensating for the lack of inductive bias toward temporal features in InstructTime. \revised{Although our experiments focus on time series classification, the framework is inherently task-agnostic and can be extended to forecasting and anomaly detection. These extensions highlight its broader applicability.} Extensive experiments conducted on multiple benchmark datasets demonstrate that the proposed methods achieve consistent and significant performance gains across diverse scenarios. Further ablation studies and analytical evaluations validated the effectiveness of each proposed component. \revised{Looking ahead, this line of work opens up promising directions toward agentic time series classification, where models can autonomously leverage external tools, memory, and iterative reasoning to perform more adaptive and context-aware decision-making.}

\section*{Acknowledgements}
This work was supported by grants from the National Key Research and Development Program of China (Grant No. 2024YFC3308200), the National Natural Science Foundation of China (No. 62502486), Anhui Provincial Natural Science Foundation (No. 2308085MG226, No. 2408085QF193), the Fundamental Research Funds for the Central Universities of China (No. WK2150110032), USTC Research Funds of the DoubleFirst-Class Initiative (No. YD2150002501).

\bibliographystyle{ACM-Reference-Format}
\bibliography{main}


\begin{thebibliography}{58}


\ifx \showCODEN    \undefined \def \showCODEN     #1{\unskip}     \fi
\ifx \showISBNx    \undefined \def \showISBNx     #1{\unskip}     \fi
\ifx \showISBNxiii \undefined \def \showISBNxiii  #1{\unskip}     \fi
\ifx \showISSN     \undefined \def \showISSN      #1{\unskip}     \fi
\ifx \showLCCN     \undefined \def \showLCCN      #1{\unskip}     \fi
\ifx \shownote     \undefined \def \shownote      #1{#1}          \fi
\ifx \showarticletitle \undefined \def \showarticletitle #1{#1}   \fi
\ifx \showURL      \undefined \def \showURL       {\relax}        \fi
\providecommand\bibfield[2]{#2}
\providecommand\bibinfo[2]{#2}
\providecommand\natexlab[1]{#1}
\providecommand\showeprint[2][]{arXiv:#2}

\bibitem[Andrzejak et~al\mbox{.}(2001)]%
        {andrzejak2001indications}
\bibfield{author}{\bibinfo{person}{Ralph~G Andrzejak}, \bibinfo{person}{Klaus
  Lehnertz}, \bibinfo{person}{Florian Mormann}, \bibinfo{person}{Christoph
  Rieke}, \bibinfo{person}{Peter David}, {and} \bibinfo{person}{Christian~E
  Elger}.} \bibinfo{year}{2001}\natexlab{}.
\newblock \showarticletitle{Indications of nonlinear deterministic and
  finite-dimensional structures in time series of brain electrical activity:
  Dependence on recording region and brain state}.
\newblock \bibinfo{journal}{\emph{Physical Review E}} \bibinfo{volume}{64},
  \bibinfo{number}{6} (\bibinfo{year}{2001}), \bibinfo{pages}{061907}.
\newblock


\bibitem[Anguita et~al\mbox{.}(2013)]%
        {anguita2013public}
\bibfield{author}{\bibinfo{person}{Davide Anguita}, \bibinfo{person}{Alessandro
  Ghio}, \bibinfo{person}{Luca Oneto}, \bibinfo{person}{Xavier Parra},
  \bibinfo{person}{Jorge~Luis Reyes-Ortiz}, {et~al\mbox{.}}}
  \bibinfo{year}{2013}\natexlab{}.
\newblock \showarticletitle{A public domain dataset for human activity
  recognition using smartphones.}. In \bibinfo{booktitle}{\emph{Esann}},
  Vol.~\bibinfo{volume}{3}. \bibinfo{pages}{3--4}.
\newblock


\bibitem[Bagnall et~al\mbox{.}(2018)]%
        {bagnall2018uea}
\bibfield{author}{\bibinfo{person}{Anthony Bagnall}, \bibinfo{person}{Hoang~Anh
  Dau}, \bibinfo{person}{Jason Lines}, \bibinfo{person}{Michael Flynn},
  \bibinfo{person}{James Large}, \bibinfo{person}{Aaron Bostrom},
  \bibinfo{person}{Paul Southam}, {and} \bibinfo{person}{Eamonn Keogh}.}
  \bibinfo{year}{2018}\natexlab{}.
\newblock \showarticletitle{The UEA multivariate time series classification
  archive, 2018}.
\newblock \bibinfo{journal}{\emph{arXiv preprint arXiv:1811.00075}}
  (\bibinfo{year}{2018}).
\newblock


\bibitem[Bengio et~al\mbox{.}(2013)]%
        {bengio2013estimating}
\bibfield{author}{\bibinfo{person}{Yoshua Bengio}, \bibinfo{person}{Nicholas
  L{\'e}onard}, {and} \bibinfo{person}{Aaron Courville}.}
  \bibinfo{year}{2013}\natexlab{}.
\newblock \showarticletitle{Estimating or propagating gradients through
  stochastic neurons for conditional computation}.
\newblock \bibinfo{journal}{\emph{arXiv preprint arXiv:1308.3432}}
  (\bibinfo{year}{2013}).
\newblock


\bibitem[Brown et~al\mbox{.}(2020)]%
        {brown2020language}
\bibfield{author}{\bibinfo{person}{Tom Brown}, \bibinfo{person}{Benjamin Mann},
  \bibinfo{person}{Nick Ryder}, \bibinfo{person}{Melanie Subbiah},
  \bibinfo{person}{Jared~D Kaplan}, \bibinfo{person}{Prafulla Dhariwal},
  \bibinfo{person}{Arvind Neelakantan}, \bibinfo{person}{Pranav Shyam},
  \bibinfo{person}{Girish Sastry}, \bibinfo{person}{Amanda Askell},
  {et~al\mbox{.}}} \bibinfo{year}{2020}\natexlab{}.
\newblock \showarticletitle{Language models are few-shot learners}.
\newblock \bibinfo{journal}{\emph{Advances in neural information processing
  systems}}  \bibinfo{volume}{33} (\bibinfo{year}{2020}),
  \bibinfo{pages}{1877--1901}.
\newblock


\bibitem[Cao et~al\mbox{.}(2024)]%
        {cao2023tempo}
\bibfield{author}{\bibinfo{person}{Defu Cao}, \bibinfo{person}{Furong Jia},
  \bibinfo{person}{Sercan~O Arik}, \bibinfo{person}{Tomas Pfister},
  \bibinfo{person}{Yixiang Zheng}, \bibinfo{person}{Wen Ye}, {and}
  \bibinfo{person}{Yan Liu}.} \bibinfo{year}{2024}\natexlab{}.
\newblock \showarticletitle{Tempo: Prompt-based generative pre-trained
  transformer for time series forecasting}. In \bibinfo{booktitle}{\emph{The
  Twelfth International Conference on Learning Representations}}.
\newblock


\bibitem[Chen et~al\mbox{.}(2025)]%
        {chen2025prospective}
\bibfield{author}{\bibinfo{person}{Jiazhen Chen}, \bibinfo{person}{Mingbin
  Feng}, {and} \bibinfo{person}{Tony~S Wirjanto}.}
  \bibinfo{year}{2025}\natexlab{}.
\newblock \showarticletitle{Prospective Multi-Graph Cohesion for Multivariate
  Time Series Anomaly Detection}. In \bibinfo{booktitle}{\emph{Proceedings of
  the Eighteenth ACM International Conference on Web Search and Data Mining}}.
  \bibinfo{pages}{98--106}.
\newblock


\bibitem[Chen et~al\mbox{.}(2023)]%
        {chen2023adversarial}
\bibfield{author}{\bibinfo{person}{Xuanhao Chen}, \bibinfo{person}{Liwei Deng},
  \bibinfo{person}{Yan Zhao}, {and} \bibinfo{person}{Kai Zheng}.}
  \bibinfo{year}{2023}\natexlab{}.
\newblock \showarticletitle{Adversarial autoencoder for unsupervised time
  series anomaly detection and interpretation}. In
  \bibinfo{booktitle}{\emph{Proceedings of the Sixteenth ACM International
  Conference on Web Search and Data Mining}}. \bibinfo{pages}{267--275}.
\newblock


\bibitem[Cheng et~al\mbox{.}(2023)]%
        {cheng2023formertime}
\bibfield{author}{\bibinfo{person}{Mingyue Cheng}, \bibinfo{person}{Qi Liu},
  \bibinfo{person}{Zhiding Liu}, \bibinfo{person}{Zhi Li},
  \bibinfo{person}{Yucong Luo}, {and} \bibinfo{person}{Enhong Chen}.}
  \bibinfo{year}{2023}\natexlab{}.
\newblock \showarticletitle{Formertime: Hierarchical multi-scale
  representations for multivariate time series classification}. In
  \bibinfo{booktitle}{\emph{Proceedings of the ACM web conference 2023}}.
  \bibinfo{pages}{1437--1445}.
\newblock


\bibitem[Cheng et~al\mbox{.}(2025a)]%
        {cheng2025cross}
\bibfield{author}{\bibinfo{person}{Mingyue Cheng}, \bibinfo{person}{Xiaoyu
  Tao}, \bibinfo{person}{Qi Liu}, \bibinfo{person}{Hao Zhang},
  \bibinfo{person}{Yiheng Chen}, {and} \bibinfo{person}{Defu Lian}.}
  \bibinfo{year}{2025}\natexlab{a}.
\newblock \showarticletitle{Cross-domain pre-training with language models for
  transferable time series representations}. In
  \bibinfo{booktitle}{\emph{Proceedings of the Eighteenth ACM International
  Conference on Web Search and Data Mining}}. \bibinfo{pages}{175--183}.
\newblock


\bibitem[Cheng et~al\mbox{.}(2026)]%
        {cheng2026timemae}
\bibfield{author}{\bibinfo{person}{Mingyue Cheng}, \bibinfo{person}{Xiaoyu
  Tao}, \bibinfo{person}{Zhiding Liu}, \bibinfo{person}{Qi Liu},
  \bibinfo{person}{Hao Zhang}, \bibinfo{person}{Rujiao Zhang}, {and}
  \bibinfo{person}{Enhong Chen}.} \bibinfo{year}{2026}\natexlab{}.
\newblock \showarticletitle{TimeMAE: Self-supervised representations of time
  series with decoupled masked autoencoders}. In
  \bibinfo{booktitle}{\emph{Proceedings of the Nineteenth ACM International
  Conference on Web Search and Data Mining}}. \bibinfo{pages}{498--508}.
\newblock


\bibitem[Cheng et~al\mbox{.}(2025b)]%
        {cheng2024convtimenet}
\bibfield{author}{\bibinfo{person}{Mingyue Cheng}, \bibinfo{person}{Jiqian
  Yang}, \bibinfo{person}{Tingyue Pan}, \bibinfo{person}{Qi Liu},
  \bibinfo{person}{Zhi Li}, {and} \bibinfo{person}{Shijin Wang}.}
  \bibinfo{year}{2025}\natexlab{b}.
\newblock \showarticletitle{Convtimenet: A deep hierarchical fully
  convolutional model for multivariate time series analysis}. In
  \bibinfo{booktitle}{\emph{Companion Proceedings of the ACM on Web Conference
  2025}}. \bibinfo{pages}{171--180}.
\newblock


\bibitem[Christ et~al\mbox{.}(2018)]%
        {christ2018time}
\bibfield{author}{\bibinfo{person}{Maximilian Christ}, \bibinfo{person}{Nils
  Braun}, \bibinfo{person}{Julius Neuffer}, {and} \bibinfo{person}{Andreas~W
  Kempa-Liehr}.} \bibinfo{year}{2018}\natexlab{}.
\newblock \showarticletitle{Time series feature extraction on basis of scalable
  hypothesis tests (tsfresh--a python package)}.
\newblock \bibinfo{journal}{\emph{Neurocomputing}}  \bibinfo{volume}{307}
  (\bibinfo{year}{2018}), \bibinfo{pages}{72--77}.
\newblock


\bibitem[Dempster et~al\mbox{.}(2021)]%
        {dempster2021minirocket}
\bibfield{author}{\bibinfo{person}{Angus Dempster}, \bibinfo{person}{Daniel~F
  Schmidt}, {and} \bibinfo{person}{Geoffrey~I Webb}.}
  \bibinfo{year}{2021}\natexlab{}.
\newblock \showarticletitle{Minirocket: A very fast (almost) deterministic
  transform for time series classification}. In
  \bibinfo{booktitle}{\emph{Proceedings of the 27th ACM SIGKDD conference on
  knowledge discovery \& data mining}}. \bibinfo{pages}{248--257}.
\newblock


\bibitem[Deng et~al\mbox{.}(2013)]%
        {deng2013time}
\bibfield{author}{\bibinfo{person}{Houtao Deng}, \bibinfo{person}{George
  Runger}, \bibinfo{person}{Eugene Tuv}, {and} \bibinfo{person}{Martyanov
  Vladimir}.} \bibinfo{year}{2013}\natexlab{}.
\newblock \showarticletitle{A time series forest for classification and feature
  extraction}.
\newblock \bibinfo{journal}{\emph{Information Sciences}}  \bibinfo{volume}{239}
  (\bibinfo{year}{2013}), \bibinfo{pages}{142--153}.
\newblock


\bibitem[Ding et~al\mbox{.}(2008)]%
        {ding2008querying}
\bibfield{author}{\bibinfo{person}{Hui Ding}, \bibinfo{person}{Goce
  Trajcevski}, \bibinfo{person}{Peter Scheuermann}, \bibinfo{person}{Xiaoyue
  Wang}, {and} \bibinfo{person}{Eamonn Keogh}.}
  \bibinfo{year}{2008}\natexlab{}.
\newblock \showarticletitle{Querying and mining of time series data:
  experimental comparison of representations and distance measures}.
\newblock \bibinfo{journal}{\emph{Proceedings of the VLDB Endowment}}
  \bibinfo{volume}{1}, \bibinfo{number}{2} (\bibinfo{year}{2008}),
  \bibinfo{pages}{1542--1552}.
\newblock


\bibitem[Eldele et~al\mbox{.}(2021)]%
        {eldele2021time}
\bibfield{author}{\bibinfo{person}{Emadeldeen Eldele}, \bibinfo{person}{Mohamed
  Ragab}, \bibinfo{person}{Zhenghua Chen}, \bibinfo{person}{Min Wu},
  \bibinfo{person}{Chee~Keong Kwoh}, \bibinfo{person}{Xiaoli Li}, {and}
  \bibinfo{person}{Cuntai Guan}.} \bibinfo{year}{2021}\natexlab{}.
\newblock \showarticletitle{Time-Series Representation Learning via Temporal
  and Contextual Contrasting}. In \bibinfo{booktitle}{\emph{Proceedings of the
  Thirtieth International Joint Conference on Artificial Intelligence}}.
  International Joint Conferences on Artificial Intelligence Organization,
  \bibinfo{pages}{2352--2359}.
\newblock


\bibitem[Gao et~al\mbox{.}(2023)]%
        {gao2023adaptive}
\bibfield{author}{\bibinfo{person}{Lina Gao}, \bibinfo{person}{Bing Liu},
  \bibinfo{person}{Ping Fu}, {and} \bibinfo{person}{Mingzhu Xu}.}
  \bibinfo{year}{2023}\natexlab{}.
\newblock \showarticletitle{Adaptive spatial tokenization transformer for
  salient object detection in optical remote sensing images}.
\newblock \bibinfo{journal}{\emph{IEEE Transactions on Geoscience and Remote
  Sensing}}  \bibinfo{volume}{61} (\bibinfo{year}{2023}),
  \bibinfo{pages}{1--15}.
\newblock


\bibitem[Grabocka et~al\mbox{.}(2014)]%
        {grabocka2014learning}
\bibfield{author}{\bibinfo{person}{Josif Grabocka}, \bibinfo{person}{Nicolas
  Schilling}, \bibinfo{person}{Martin Wistuba}, {and} \bibinfo{person}{Lars
  Schmidt-Thieme}.} \bibinfo{year}{2014}\natexlab{}.
\newblock \showarticletitle{Learning time-series shapelets}. In
  \bibinfo{booktitle}{\emph{Proceedings of the 20th ACM SIGKDD international
  conference on Knowledge discovery and data mining}}.
  \bibinfo{pages}{392--401}.
\newblock


\bibitem[Gray(1984)]%
        {gray1984vector}
\bibfield{author}{\bibinfo{person}{Robert Gray}.}
  \bibinfo{year}{1984}\natexlab{}.
\newblock \showarticletitle{Vector quantization}.
\newblock \bibinfo{journal}{\emph{IEEE Assp Magazine}} \bibinfo{volume}{1},
  \bibinfo{number}{2} (\bibinfo{year}{1984}), \bibinfo{pages}{4--29}.
\newblock


\bibitem[Gupta et~al\mbox{.}(2020)]%
        {gupta2020approaches}
\bibfield{author}{\bibinfo{person}{Ashish Gupta}, \bibinfo{person}{Hari~Prabhat
  Gupta}, \bibinfo{person}{Bhaskar Biswas}, {and} \bibinfo{person}{Tanima
  Dutta}.} \bibinfo{year}{2020}\natexlab{}.
\newblock \showarticletitle{Approaches and applications of early classification
  of time series: A review}.
\newblock \bibinfo{journal}{\emph{IEEE Transactions on Artificial
  Intelligence}} \bibinfo{volume}{1}, \bibinfo{number}{1}
  (\bibinfo{year}{2020}), \bibinfo{pages}{47--61}.
\newblock


\bibitem[Han et~al\mbox{.}(2021)]%
        {han2021dynamic}
\bibfield{author}{\bibinfo{person}{Liangzhe Han}, \bibinfo{person}{Bowen Du},
  \bibinfo{person}{Leilei Sun}, \bibinfo{person}{Yanjie Fu},
  \bibinfo{person}{Yisheng Lv}, {and} \bibinfo{person}{Hui Xiong}.}
  \bibinfo{year}{2021}\natexlab{}.
\newblock \showarticletitle{Dynamic and multi-faceted spatio-temporal deep
  learning for traffic speed forecasting}. In
  \bibinfo{booktitle}{\emph{Proceedings of the 27th ACM SIGKDD conference on
  knowledge discovery \& data mining}}. \bibinfo{pages}{547--555}.
\newblock


\bibitem[Hinton(2014)]%
        {hinton2015distilling}
\bibfield{author}{\bibinfo{person}{G Hinton}.} \bibinfo{year}{2014}\natexlab{}.
\newblock \showarticletitle{Distilling the Knowledge in a Neural Network}. In
  \bibinfo{booktitle}{\emph{Deep Learning and Representation Learning Workshop
  in Conjunction with NIPS}}.
\newblock


\bibitem[Ismail~Fawaz et~al\mbox{.}(2020)]%
        {ismail2020inceptiontime}
\bibfield{author}{\bibinfo{person}{Hassan Ismail~Fawaz},
  \bibinfo{person}{Benjamin Lucas}, \bibinfo{person}{Germain Forestier},
  \bibinfo{person}{Charlotte Pelletier}, \bibinfo{person}{Daniel~F Schmidt},
  \bibinfo{person}{Jonathan Weber}, \bibinfo{person}{Geoffrey~I Webb},
  \bibinfo{person}{Lhassane Idoumghar}, \bibinfo{person}{Pierre-Alain Muller},
  {and} \bibinfo{person}{Fran{\c{c}}ois Petitjean}.}
  \bibinfo{year}{2020}\natexlab{}.
\newblock \showarticletitle{Inceptiontime: Finding alexnet for time series
  classification}.
\newblock \bibinfo{journal}{\emph{Data Mining and Knowledge Discovery}}
  \bibinfo{volume}{34}, \bibinfo{number}{6} (\bibinfo{year}{2020}),
  \bibinfo{pages}{1936--1962}.
\newblock


\bibitem[Kenton and Toutanova(2019)]%
        {devlin2018bert}
\bibfield{author}{\bibinfo{person}{Jacob Devlin Ming-Wei~Chang Kenton} {and}
  \bibinfo{person}{Lee~Kristina Toutanova}.} \bibinfo{year}{2019}\natexlab{}.
\newblock \showarticletitle{BERT: Pre-training of Deep Bidirectional
  Transformers for Language Understanding}. In
  \bibinfo{booktitle}{\emph{Proceedings of NAACL-HLT}}.
  \bibinfo{pages}{4171--4186}.
\newblock


\bibitem[Lewis et~al\mbox{.}(2020)]%
        {lewis2019bart}
\bibfield{author}{\bibinfo{person}{Mike Lewis}, \bibinfo{person}{Yinhan Liu},
  \bibinfo{person}{Naman Goyal}, \bibinfo{person}{Marjan Ghazvininejad},
  \bibinfo{person}{Abdelrahman Mohamed}, \bibinfo{person}{Omer Levy},
  \bibinfo{person}{Veselin Stoyanov}, {and} \bibinfo{person}{Luke
  Zettlemoyer}.} \bibinfo{year}{2020}\natexlab{}.
\newblock \showarticletitle{BART: Denoising sequence-to-sequence pre-training
  for natural language generation, translation, and comprehension}. In
  \bibinfo{booktitle}{\emph{Proceedings of the 58th annual meeting of the
  association for computational linguistics}}. \bibinfo{pages}{7871--7880}.
\newblock


\bibitem[Li et~al\mbox{.}(2025)]%
        {li2025incomplete}
\bibfield{author}{\bibinfo{person}{Xiaocui Li}, \bibinfo{person}{Guoliang Li},
  \bibinfo{person}{Xinyu Zhang}, \bibinfo{person}{Yangtao Wang},
  \bibinfo{person}{Qingyu Shi}, {and} \bibinfo{person}{Wei Liang}.}
  \bibinfo{year}{2025}\natexlab{}.
\newblock \showarticletitle{Incomplete Multi-view Clustering via Local
  Reasoning and Correlation Analysis}. In \bibinfo{booktitle}{\emph{Proceedings
  of the Eighteenth ACM International Conference on Web Search and Data
  Mining}}. \bibinfo{pages}{484--492}.
\newblock


\bibitem[Lin et~al\mbox{.}(2007)]%
        {lin2007experiencing}
\bibfield{author}{\bibinfo{person}{Jessica Lin}, \bibinfo{person}{Eamonn
  Keogh}, \bibinfo{person}{Li Wei}, {and} \bibinfo{person}{Stefano Lonardi}.}
  \bibinfo{year}{2007}\natexlab{}.
\newblock \showarticletitle{Experiencing SAX: a novel symbolic representation
  of time series}.
\newblock \bibinfo{journal}{\emph{Data Mining and knowledge discovery}}
  \bibinfo{volume}{15} (\bibinfo{year}{2007}), \bibinfo{pages}{107--144}.
\newblock


\bibitem[Lin et~al\mbox{.}(2012)]%
        {lin2012rotation}
\bibfield{author}{\bibinfo{person}{Jessica Lin}, \bibinfo{person}{Rohan Khade},
  {and} \bibinfo{person}{Yuan Li}.} \bibinfo{year}{2012}\natexlab{}.
\newblock \showarticletitle{Rotation-invariant similarity in time series using
  bag-of-patterns representation}.
\newblock \bibinfo{journal}{\emph{Journal of Intelligent Information Systems}}
  \bibinfo{volume}{39} (\bibinfo{year}{2012}), \bibinfo{pages}{287--315}.
\newblock


\bibitem[Liu et~al\mbox{.}(2021a)]%
        {liu2021pay}
\bibfield{author}{\bibinfo{person}{Hanxiao Liu}, \bibinfo{person}{Zihang Dai},
  \bibinfo{person}{David So}, {and} \bibinfo{person}{Quoc~V Le}.}
  \bibinfo{year}{2021}\natexlab{a}.
\newblock \showarticletitle{Pay attention to mlps}.
\newblock \bibinfo{journal}{\emph{Advances in neural information processing
  systems}}  \bibinfo{volume}{34} (\bibinfo{year}{2021}),
  \bibinfo{pages}{9204--9215}.
\newblock


\bibitem[Liu et~al\mbox{.}(2021b)]%
        {liu2021gated}
\bibfield{author}{\bibinfo{person}{Minghao Liu}, \bibinfo{person}{Shengqi Ren},
  \bibinfo{person}{Siyuan Ma}, \bibinfo{person}{Jiahui Jiao},
  \bibinfo{person}{Yizhou Chen}, \bibinfo{person}{Zhiguang Wang}, {and}
  \bibinfo{person}{Wei Song}.} \bibinfo{year}{2021}\natexlab{b}.
\newblock \showarticletitle{Gated transformer networks for multivariate time
  series classification}.
\newblock \bibinfo{journal}{\emph{arXiv preprint arXiv:2103.14438}}
  (\bibinfo{year}{2021}).
\newblock


\bibitem[Liu et~al\mbox{.}(2019)]%
        {liu2019ekt}
\bibfield{author}{\bibinfo{person}{Qi Liu}, \bibinfo{person}{Zhenya Huang},
  \bibinfo{person}{Yu Yin}, \bibinfo{person}{Enhong Chen}, \bibinfo{person}{Hui
  Xiong}, \bibinfo{person}{Yu Su}, {and} \bibinfo{person}{Guoping Hu}.}
  \bibinfo{year}{2019}\natexlab{}.
\newblock \showarticletitle{Ekt: Exercise-aware knowledge tracing for student
  performance prediction}.
\newblock \bibinfo{journal}{\emph{IEEE Transactions on Knowledge and Data
  Engineering}} \bibinfo{volume}{33}, \bibinfo{number}{1}
  (\bibinfo{year}{2019}), \bibinfo{pages}{100--115}.
\newblock


\bibitem[Liu et~al\mbox{.}(2023)]%
        {liu2024adaptive}
\bibfield{author}{\bibinfo{person}{Zhiding Liu}, \bibinfo{person}{Mingyue
  Cheng}, \bibinfo{person}{Zhi Li}, \bibinfo{person}{Zhenya Huang},
  \bibinfo{person}{Qi Liu}, \bibinfo{person}{Yanhu Xie}, {and}
  \bibinfo{person}{Enhong Chen}.} \bibinfo{year}{2023}\natexlab{}.
\newblock \showarticletitle{Adaptive normalization for non-stationary time
  series forecasting: A temporal slice perspective}.
\newblock \bibinfo{journal}{\emph{Advances in Neural Information Processing
  Systems}}  \bibinfo{volume}{36} (\bibinfo{year}{2023}),
  \bibinfo{pages}{14273--14292}.
\newblock


\bibitem[Liu et~al\mbox{.}(2024)]%
        {liu2024generative}
\bibfield{author}{\bibinfo{person}{Zhiding Liu}, \bibinfo{person}{Jiqian Yang},
  \bibinfo{person}{Mingyue Cheng}, \bibinfo{person}{Yucong Luo}, {and}
  \bibinfo{person}{Zhi Li}.} \bibinfo{year}{2024}\natexlab{}.
\newblock \showarticletitle{Generative pretrained hierarchical transformer for
  time series forecasting}. In \bibinfo{booktitle}{\emph{Proceedings of the
  30th ACM SIGKDD conference on knowledge discovery and data mining}}.
  \bibinfo{pages}{2003--2013}.
\newblock


\bibitem[Middlehurst et~al\mbox{.}(2024)]%
        {middlehurst2023bake}
\bibfield{author}{\bibinfo{person}{Matthew Middlehurst},
  \bibinfo{person}{Patrick Sch{\"a}fer}, {and} \bibinfo{person}{Anthony
  Bagnall}.} \bibinfo{year}{2024}\natexlab{}.
\newblock \showarticletitle{Bake off redux: a review and experimental
  evaluation of recent time series classification algorithms}.
\newblock \bibinfo{journal}{\emph{Data Mining and Knowledge Discovery}}
  \bibinfo{volume}{38}, \bibinfo{number}{4} (\bibinfo{year}{2024}),
  \bibinfo{pages}{1958--2031}.
\newblock


\bibitem[Mikolov et~al\mbox{.}(2013)]%
        {mikolov2013efficient}
\bibfield{author}{\bibinfo{person}{Tomas Mikolov}, \bibinfo{person}{Kai Chen},
  \bibinfo{person}{Greg Corrado}, {and} \bibinfo{person}{Jeffrey Dean}.}
  \bibinfo{year}{2013}\natexlab{}.
\newblock \showarticletitle{Efficient estimation of word representations in
  vector space}. In \bibinfo{booktitle}{\emph{International Conference on
  Learning Representations}}.
\newblock


\bibitem[Ouyang et~al\mbox{.}(2022)]%
        {ouyang2022training}
\bibfield{author}{\bibinfo{person}{Long Ouyang}, \bibinfo{person}{Jeffrey Wu},
  \bibinfo{person}{Xu Jiang}, \bibinfo{person}{Diogo Almeida},
  \bibinfo{person}{Carroll Wainwright}, \bibinfo{person}{Pamela Mishkin},
  \bibinfo{person}{Chong Zhang}, \bibinfo{person}{Sandhini Agarwal},
  \bibinfo{person}{Katarina Slama}, \bibinfo{person}{Alex Ray},
  {et~al\mbox{.}}} \bibinfo{year}{2022}\natexlab{}.
\newblock \showarticletitle{Training language models to follow instructions
  with human feedback}.
\newblock \bibinfo{journal}{\emph{Advances in Neural Information Processing
  Systems}}  \bibinfo{volume}{35} (\bibinfo{year}{2022}),
  \bibinfo{pages}{27730--27744}.
\newblock


\bibitem[Petitjean et~al\mbox{.}(2014)]%
        {petitjean2014dynamic}
\bibfield{author}{\bibinfo{person}{Fran{\c{c}}ois Petitjean},
  \bibinfo{person}{Germain Forestier}, \bibinfo{person}{Geoffrey~I Webb},
  \bibinfo{person}{Ann~E Nicholson}, \bibinfo{person}{Yanping Chen}, {and}
  \bibinfo{person}{Eamonn Keogh}.} \bibinfo{year}{2014}\natexlab{}.
\newblock \showarticletitle{Dynamic time warping averaging of time series
  allows faster and more accurate classification}. In
  \bibinfo{booktitle}{\emph{2014 IEEE international conference on data
  mining}}. IEEE, \bibinfo{pages}{470--479}.
\newblock


\bibitem[Radford et~al\mbox{.}(2018)]%
        {radford2018improving}
\bibfield{author}{\bibinfo{person}{Alec Radford}, \bibinfo{person}{Karthik
  Narasimhan}, \bibinfo{person}{Tim Salimans}, \bibinfo{person}{Ilya
  Sutskever}, {et~al\mbox{.}}} \bibinfo{year}{2018}\natexlab{}.
\newblock \showarticletitle{Improving language understanding by generative
  pre-training}.
\newblock  (\bibinfo{year}{2018}).
\newblock


\bibitem[Radford et~al\mbox{.}(2019)]%
        {radford2019language}
\bibfield{author}{\bibinfo{person}{Alec Radford}, \bibinfo{person}{Jeffrey Wu},
  \bibinfo{person}{Rewon Child}, \bibinfo{person}{David Luan},
  \bibinfo{person}{Dario Amodei}, \bibinfo{person}{Ilya Sutskever},
  {et~al\mbox{.}}} \bibinfo{year}{2019}\natexlab{}.
\newblock \showarticletitle{Language models are unsupervised multitask
  learners}.
\newblock \bibinfo{journal}{\emph{OpenAI blog}} \bibinfo{volume}{1},
  \bibinfo{number}{8} (\bibinfo{year}{2019}), \bibinfo{pages}{9}.
\newblock


\bibitem[Rendle(2010)]%
        {rendle2010factorization}
\bibfield{author}{\bibinfo{person}{Steffen Rendle}.}
  \bibinfo{year}{2010}\natexlab{}.
\newblock \showarticletitle{Factorization machines}. In
  \bibinfo{booktitle}{\emph{2010 IEEE International conference on data
  mining}}. IEEE, \bibinfo{pages}{995--1000}.
\newblock


\bibitem[Senin and Malinchik(2013)]%
        {senin2013sax}
\bibfield{author}{\bibinfo{person}{Pavel Senin} {and} \bibinfo{person}{Sergey
  Malinchik}.} \bibinfo{year}{2013}\natexlab{}.
\newblock \showarticletitle{Sax-vsm: Interpretable time series classification
  using sax and vector space model}. In \bibinfo{booktitle}{\emph{2013 IEEE
  13th international conference on data mining}}. IEEE,
  \bibinfo{pages}{1175--1180}.
\newblock


\bibitem[Shifaz et~al\mbox{.}(2020)]%
        {shifaz2020ts}
\bibfield{author}{\bibinfo{person}{Ahmed Shifaz}, \bibinfo{person}{Charlotte
  Pelletier}, \bibinfo{person}{Fran{\c{c}}ois Petitjean}, {and}
  \bibinfo{person}{Geoffrey~I Webb}.} \bibinfo{year}{2020}\natexlab{}.
\newblock \showarticletitle{TS-CHIEF: a scalable and accurate forest algorithm
  for time series classification}.
\newblock \bibinfo{journal}{\emph{Data Mining and Knowledge Discovery}}
  \bibinfo{volume}{34}, \bibinfo{number}{3} (\bibinfo{year}{2020}),
  \bibinfo{pages}{742--775}.
\newblock


\bibitem[Song et~al\mbox{.}(2025)]%
        {song2025comprehensive}
\bibfield{author}{\bibinfo{person}{Xuemeng Song}, \bibinfo{person}{Haoqiang
  Lin}, \bibinfo{person}{Haokun Wen}, \bibinfo{person}{Bohan Hou},
  \bibinfo{person}{Mingzhu Xu}, {and} \bibinfo{person}{Liqiang Nie}.}
  \bibinfo{year}{2025}\natexlab{}.
\newblock \showarticletitle{A comprehensive survey on composed image
  retrieval}.
\newblock \bibinfo{journal}{\emph{ACM Transactions on Information Systems}}
  \bibinfo{volume}{44}, \bibinfo{number}{1} (\bibinfo{year}{2025}),
  \bibinfo{pages}{1--54}.
\newblock


\bibitem[Tao et~al\mbox{.}(2026)]%
        {tao2026memcast}
\bibfield{author}{\bibinfo{person}{Xiaoyu Tao}, \bibinfo{person}{Mingyue
  Cheng}, \bibinfo{person}{Ze Guo}, \bibinfo{person}{Shuo Yu},
  \bibinfo{person}{Yaguo Liu}, \bibinfo{person}{Qi Liu}, {and}
  \bibinfo{person}{Shijin Wang}.} \bibinfo{year}{2026}\natexlab{}.
\newblock \showarticletitle{MemCast: Memory-Driven Time Series Forecasting with
  Experience-Conditioned Reasoning}.
\newblock \bibinfo{journal}{\emph{arXiv preprint arXiv:2602.03164}}
  (\bibinfo{year}{2026}).
\newblock


\bibitem[Tao et~al\mbox{.}(2024)]%
        {tao2024hierarchical}
\bibfield{author}{\bibinfo{person}{Xiaoyu Tao}, \bibinfo{person}{Tingyue Pan},
  \bibinfo{person}{Mingyue Cheng}, \bibinfo{person}{Yucong Luo},
  \bibinfo{person}{Qi Liu}, {and} \bibinfo{person}{Enhong Chen}.}
  \bibinfo{year}{2024}\natexlab{}.
\newblock \showarticletitle{Hierarchical multimodal llms with semantic space
  alignment for enhanced time series classification}.
\newblock \bibinfo{journal}{\emph{ACM Transactions on Intelligent Systems and
  Technology}} (\bibinfo{year}{2024}).
\newblock


\bibitem[van~den Oord et~al\mbox{.}(2016)]%
        {oord2016wavenet}
\bibfield{author}{\bibinfo{person}{A{\"a}ron van~den Oord},
  \bibinfo{person}{Sander Dieleman}, \bibinfo{person}{Heiga Zen},
  \bibinfo{person}{Karen Simonyan}, \bibinfo{person}{Oriol Vinyals},
  \bibinfo{person}{Alex Graves}, \bibinfo{person}{Nal Kalchbrenner},
  \bibinfo{person}{Andrew Senior}, {and} \bibinfo{person}{Koray Kavukcuoglu}.}
  \bibinfo{year}{2016}\natexlab{}.
\newblock \showarticletitle{WaveNet: A Generative Model for Raw Audio}. In
  \bibinfo{booktitle}{\emph{Proc. SSW 2016}}. \bibinfo{pages}{125--125}.
\newblock


\bibitem[Van Den~Oord et~al\mbox{.}(2017)]%
        {van2017neural}
\bibfield{author}{\bibinfo{person}{Aaron Van Den~Oord}, \bibinfo{person}{Oriol
  Vinyals}, {et~al\mbox{.}}} \bibinfo{year}{2017}\natexlab{}.
\newblock \showarticletitle{Neural discrete representation learning}.
\newblock \bibinfo{journal}{\emph{Advances in neural information processing
  systems}}  \bibinfo{volume}{30} (\bibinfo{year}{2017}).
\newblock


\bibitem[Wang et~al\mbox{.}(2024)]%
        {wang2024tabletime}
\bibfield{author}{\bibinfo{person}{Jiahao Wang}, \bibinfo{person}{Mingyue
  Cheng}, \bibinfo{person}{Qingyang Mao}, \bibinfo{person}{Yitong Zhou},
  \bibinfo{person}{Feiyang Xu}, {and} \bibinfo{person}{Xin Li}.}
  \bibinfo{year}{2024}\natexlab{}.
\newblock \showarticletitle{TableTime: Reformulating Time Series Classification
  as Training-Free Table Understanding with Large Language Models}. In
  \bibinfo{booktitle}{\emph{Proceedings of the 34th ACM International
  Conference on Information and Knowledge Management}}.
  \bibinfo{pages}{3009–3019}.
\newblock


\bibitem[Wang(2024)]%
        {wang2024neuralreconciler}
\bibfield{author}{\bibinfo{person}{Shiyu Wang}.}
  \bibinfo{year}{2024}\natexlab{}.
\newblock \showarticletitle{Neuralreconciler for hierarchical time series
  forecasting}. In \bibinfo{booktitle}{\emph{Proceedings of the 17th ACM
  International Conference on Web Search and Data Mining}}.
  \bibinfo{pages}{731--739}.
\newblock


\bibitem[Wi et~al\mbox{.}(2024)]%
        {wi2024continuous}
\bibfield{author}{\bibinfo{person}{Hyowon Wi}, \bibinfo{person}{Yehjin Shin},
  {and} \bibinfo{person}{Noseong Park}.} \bibinfo{year}{2024}\natexlab{}.
\newblock \showarticletitle{Continuous-time autoencoders for regular and
  irregular time series imputation}. In \bibinfo{booktitle}{\emph{Proceedings
  of the 17th ACM International Conference on Web Search and Data Mining}}.
  \bibinfo{pages}{826--835}.
\newblock


\bibitem[Ye and Keogh(2009)]%
        {ye2009time}
\bibfield{author}{\bibinfo{person}{Lexiang Ye} {and} \bibinfo{person}{Eamonn
  Keogh}.} \bibinfo{year}{2009}\natexlab{}.
\newblock \showarticletitle{Time series shapelets: a new primitive for data
  mining}. In \bibinfo{booktitle}{\emph{Proceedings of the 15th ACM SIGKDD
  international conference on Knowledge discovery and data mining}}.
  \bibinfo{pages}{947--956}.
\newblock


\bibitem[Yuan et~al\mbox{.}(2025)]%
        {yuan2025watermarking}
\bibfield{author}{\bibinfo{person}{Wei Yuan}, \bibinfo{person}{Chaoqun Yang},
  \bibinfo{person}{Yu Xing}, \bibinfo{person}{Tong Chen},
  \bibinfo{person}{Nguyen Quoc~Viet Hung}, {and} \bibinfo{person}{Hongzhi
  Yin}.} \bibinfo{year}{2025}\natexlab{}.
\newblock \showarticletitle{Watermarking Large Language Model-based Time Series
  Forecasting}.
\newblock \bibinfo{journal}{\emph{arXiv preprint arXiv:2507.20762}}
  (\bibinfo{year}{2025}).
\newblock


\bibitem[Zerveas et~al\mbox{.}(2021)]%
        {zerveas2021transformer}
\bibfield{author}{\bibinfo{person}{George Zerveas}, \bibinfo{person}{Srideepika
  Jayaraman}, \bibinfo{person}{Dhaval Patel}, \bibinfo{person}{Anuradha
  Bhamidipaty}, {and} \bibinfo{person}{Carsten Eickhoff}.}
  \bibinfo{year}{2021}\natexlab{}.
\newblock \showarticletitle{A transformer-based framework for multivariate time
  series representation learning}. In \bibinfo{booktitle}{\emph{Proceedings of
  the 27th ACM SIGKDD conference on knowledge discovery \& data mining}}.
  \bibinfo{pages}{2114--2124}.
\newblock


\bibitem[Zhang et~al\mbox{.}(2020)]%
        {zhang2020tapnet}
\bibfield{author}{\bibinfo{person}{Xuchao Zhang}, \bibinfo{person}{Yifeng Gao},
  \bibinfo{person}{Jessica Lin}, {and} \bibinfo{person}{Chang-Tien Lu}.}
  \bibinfo{year}{2020}\natexlab{}.
\newblock \showarticletitle{Tapnet: Multivariate time series classification
  with attentional prototypical network}. In
  \bibinfo{booktitle}{\emph{Proceedings of the AAAI Conference on Artificial
  Intelligence}}, Vol.~\bibinfo{volume}{34}. \bibinfo{pages}{6845--6852}.
\newblock


\bibitem[Zheng et~al\mbox{.}(2014)]%
        {zheng2014time}
\bibfield{author}{\bibinfo{person}{Yi Zheng}, \bibinfo{person}{Qi Liu},
  \bibinfo{person}{Enhong Chen}, \bibinfo{person}{Yong Ge}, {and}
  \bibinfo{person}{J~Leon Zhao}.} \bibinfo{year}{2014}\natexlab{}.
\newblock \showarticletitle{Time series classification using multi-channels
  deep convolutional neural networks}. In
  \bibinfo{booktitle}{\emph{International conference on web-age information
  management}}. Springer, \bibinfo{pages}{298--310}.
\newblock


\bibitem[Zhong et~al\mbox{.}(2026)]%
        {zhong2026logicgate}
\bibfield{author}{\bibinfo{person}{Jie Zhong}, \bibinfo{person}{Tong Chen},
  \bibinfo{person}{Wei Yuan}, \bibinfo{person}{Lizhen Cui}, {and}
  \bibinfo{person}{Hongzhi Yin}.} \bibinfo{year}{2026}\natexlab{}.
\newblock \showarticletitle{LogicGate: Adaptive Rule-Based Modeling of
  Exogenous Effects for Time Series Forecasting}. In
  \bibinfo{booktitle}{\emph{International Conference on Database Systems for
  Advanced Applications}}. Springer, \bibinfo{pages}{239--255}.
\newblock


\bibitem[Zhou et~al\mbox{.}(2021)]%
        {zhou2021informer}
\bibfield{author}{\bibinfo{person}{Haoyi Zhou}, \bibinfo{person}{Shanghang
  Zhang}, \bibinfo{person}{Jieqi Peng}, \bibinfo{person}{Shuai Zhang},
  \bibinfo{person}{Jianxin Li}, \bibinfo{person}{Hui Xiong}, {and}
  \bibinfo{person}{Wancai Zhang}.} \bibinfo{year}{2021}\natexlab{}.
\newblock \showarticletitle{Informer: Beyond efficient transformer for long
  sequence time-series forecasting}. In \bibinfo{booktitle}{\emph{Proceedings
  of the AAAI conference on artificial intelligence}},
  Vol.~\bibinfo{volume}{35}. \bibinfo{pages}{11106--11115}.
\newblock


\end{thebibliography}

\appendix

\end{document}